\documentclass[10pt]{article}

\usepackage[margin=1in]{geometry}

\usepackage[utf8]{inputenc}
\usepackage[T1]{fontenc}
\usepackage{hyperref}
\usepackage{url}
\usepackage[numbers]{natbib}
\usepackage{booktabs}
\usepackage{amsfonts}
\usepackage{nicefrac}
\usepackage{microtype}
\usepackage[dvipsnames]{xcolor}
\usepackage{enumitem}
\usepackage{subcaption}

\usepackage{graphicx}
\usepackage{tikz}
\usetikzlibrary{shapes.geometric,positioning,decorations.pathreplacing}
\usepackage{mathtools}
\usepackage{amsthm}
\usepackage{thmtools,thm-restate}
\usepackage{aligned-overset}
\usepackage{enumitem}
\usepackage{caption}
\usepackage{wrapfig}
\usepackage{bm}
\usepackage{algorithm}
\usepackage{algorithmic}
\usepackage{listings}
\usepackage{float}

\DeclareMathOperator*{\argmin}{arg\,min}

\definecolor{codegreen}{rgb}{0,0.6,0}
\definecolor{codegray}{rgb}{0.5,0.5,0.5}
\definecolor{codepurple}{rgb}{0.58,0,0.82}
\definecolor{backcolour}{rgb}{0.95,0.95,0.92}

\lstdefinestyle{mystyle}{
    backgroundcolor=\color{backcolour},   
    commentstyle=\color{codegreen},
    keywordstyle=\color{magenta},
    numberstyle=\tiny\color{codegray},
    stringstyle=\color{codepurple},
    basicstyle=\ttfamily\scriptsize,
    breakatwhitespace=false,         
    breaklines=true,                 
    captionpos=b,                    
    keepspaces=true,                 
    numbers=left,
    numbersep=3pt,                  
    showspaces=false,                
    showstringspaces=false,
    showtabs=false,                  
    tabsize=2,
    aboveskip=3pt,
    belowskip=3pt
}

\title{CRUMB: Efficient Prior Fitted Network Inference\\
via Distributionally Matched Context Batching}

\author{
  Jamie Heredge\thanks{Equal contribution. Email: \texttt{\{jamie.heredge, mattia.villani\}@jpmchase.com.}} 
  \quad
  Mattia J. Villani$^*$
  \quad
  Pranav Deshpande \quad
  Akshay Seshadri \quad
  \\\quad
  {Niraj Kumar}\thanks{Principal Investigator. Email: \texttt{niraj.x7.kumar@jpmchase.com}}\\ \\
  Global Technology Applied Research,\\ JPMorganChase, New York, NY 10001, USA\\
}

\date{}

\begin{document}
\maketitle

\begin{abstract}
Prior-fitted networks (PFNs) are a promising class of tabular foundation models that perform in-context learning, whereby the entire labelled training set is supplied as context, and predictions for test queries are produced in a single forward pass. However, the quadratically scaling self-attention mechanism in many PFN architectures makes inference prohibitive for very large training datasets. We propose \textbf{CRUMB} (\textbf{C}lustered \textbf{R}etrieval \textbf{U}sing \textbf{M}inimised-MMD \textbf{B}atching), a three-stage inference wrapper that (i)~clusters the test queries, (ii)~selects a small, distributionally matched training subset for each cluster by greedily minimising the maximum mean discrepancy (MMD), and (iii)~runs exact PFN inference on each reduced-context batch. CRUMB is architecture-agnostic and requires no retraining. On the 51-dataset TabArena benchmark, evaluated across three PFN architectures (TabPFNv2, TabICLv1, TabICLv2), we show that CRUMB outperforms similar state-of-the-art context selection strategies. We also show that CRUMB is resilient to covariate drift, as the MMD-minimisation step naturally helps align the training context distribution to match the current test batch distributions. 
\end{abstract}

\section{Introduction}
\label{sec:intro}

Prior-fitted networks (PFNs) ~\cite{hollmann2023original2023, hollmann2023tabpfntransformersolvessmall} are an increasingly popular \cite{van2024tabular} class of tabular foundation models that solve supervised learning tasks via in-context learning: the entire labelled training set is provided as context, and the model produces predictions for test points in a single forward pass. 
These PFN models have achieved success on various datasets, even outperforming gradient boosting decision tree methods such as CatBoost \cite{prokhorenkova2019catboostunbiasedboostingcategorical} and XGBoost \cite{Chen_2016}, which had previously been considered some of the most competitive models for tabular dataset classification tasks. 
One key problem is that the reported PFN victories over current state of the art methods typically only involve small datasets \cite{hollmann2023tabpfntransformersolvessmall, ye2025closer, cheng2025realistic}. 
This is partly due to the attention layers scaling quadratically with the size of the training dataset which makes the time and memory costs for PFN models prohibitive for large dataset sizes.

Multiple different methods have been suggested to overcome this issue. 
Examples of possible solutions include architectural changes to the PFN itself, such as switching to linear attention, as found in TabFlex \cite{zeng2025tabflexscalingtabularlearning}. 
There are also techniques focused on fine-tuning model weights for a specific dataset \cite{thomas2024retrievalfinetuningincontext} as well as context-tuning \cite{feuer2024tunetablescontextoptimizationscalable} where the training points themselves are variationally adapted. In this work we will not focus on architectural adjustments, instead restricting our investigation to TabPFNv2, TabICLv1 and TabICLv2.

A natural remedy is \emph{context selection}: replace the full training set
$\mathcal{D}_{\text{train}}$ with a much smaller subset $\mathcal{S} \subset \mathcal{D}_{\text{train}}$, chosen so that predictions are minimally affected. Uniform subsampling is the simplest option, although it effectively randomly discards information in exchange for improving speed. Query-dependent methods such as $k$-nearest-neighbour ($k$NN) retrieval can improve accuracy by tailoring the context to each test point, but they sacrifice the ability to batch test queries: because every test point receives a different context, each requires a separate forward pass, resulting in $T$ independent PFN evaluations rather than a small number of batched calls. 

\begin{figure}[t]
\centering
\resizebox{\textwidth}{!}{%
\begin{tikzpicture}[
    >=stealth,
    font=\small,
    databox/.style={draw=black!70, fill=blue!6, rounded corners=3pt,
                    minimum height=0.8cm, minimum width=1.6cm, align=center,
                    line width=0.6pt},
    procbox/.style={draw=black!70, rounded corners=3pt,
                    minimum height=0.8cm, minimum width=1.8cm, align=center,
                    line width=0.6pt},
    pfnbox/.style={draw=black!80, fill=green!8, rounded corners=4pt,
                   minimum height=0.85cm, minimum width=1.6cm, align=center,
                   line width=0.7pt},
    outbox/.style={draw=black!70, fill=orange!8, rounded corners=4pt,
                   minimum height=0.7cm, minimum width=1.4cm, align=center,
                   line width=0.6pt},
    arr/.style={->, thick, color=black!60},
    darr/.style={->, thick, color=violet!50, dashed},
    brack/.style={decorate, decoration={brace, amplitude=4pt, mirror}},
    lbl/.style={font=\tiny, text=black!55},
]

\node[databox, fill=blue!8, minimum width=1.4cm] (train) at (0, -0.8) 
    {\textbf{Train}\\[-2pt]{\tiny $\mathcal{D}_{\text{train}}$, $N$ pts}};

\node[databox, fill=red!8, minimum width=1.4cm] (test) at (0, -1.8) 
    {\textbf{Test}\\[-2pt]{\tiny $\mathcal{D}_{\text{test}}$, $T$ pts}};

\node[procbox, fill=red!6, minimum width=2cm] (cluster) at (3.2, -1.8) 
    {\textbf{Stage 1}\\[-2pt]{\tiny $k$-means on $\mathcal{D}_{\text{test}}$}};
\draw[arr] (test.east) -- (cluster.west);

\node[draw=red!50, fill=red!4, rounded corners=2pt, 
      minimum height=0.5cm, minimum width=1cm, font=\tiny, align=center]
      (c1) at (5.8, -0.7) {$C_1$};
\node[draw=red!50, fill=red!4, rounded corners=2pt, 
      minimum height=0.5cm, minimum width=1cm, font=\tiny, align=center]
      (c2) at (5.8, -1.8) {$C_2$};
\node[draw=red!50, fill=red!4, rounded corners=2pt, 
      minimum height=0.5cm, minimum width=1cm, font=\tiny, align=center]
      (ck) at (5.8, -2.9) {$C_K$};

\draw[arr, color=red!45] (cluster.east) -- ++(0.6,0) |- (c1.west);
\draw[arr, color=red!45] (cluster.east) -- (c2.west);
\draw[arr, color=red!45] (cluster.east) -- ++(0.6,0) |- (ck.west);

\node[font=\small, text=black!40] at (5.8, -2.25) {$\vdots$};

\node[procbox, fill=cyan!10, minimum width=2cm, minimum height=0.5cm] 
      (mmd1) at (8.2, -0.7) {{\tiny MMD Herding}};
\node[procbox, fill=cyan!10, minimum width=2cm, minimum height=0.5cm] 
      (mmd2) at (8.2, -1.8) {{\tiny MMD Herding}};
\node[procbox, fill=cyan!10, minimum width=2cm, minimum height=0.5cm] 
      (mmdk) at (8.2, -2.9) {{\tiny MMD Herding}};

\node[font=\small, text=black!40] at (8.2, -2.25) {$\vdots$};

\draw[arr] (c1.east) -- (mmd1.west);
\draw[arr] (c2.east) -- (mmd2.west);
\draw[arr] (ck.east) -- (mmdk.west);

\node[font=\tiny, text=cyan!50!black, above=0.08cm of mmd1] {\textbf{Stage 2}};

\draw[darr, color=blue!40] (train.east) -| (4.0,0) -| (mmd1.north);

\node[draw=blue!45, fill=blue!4, rounded corners=2pt,
      minimum height=0.5cm, minimum width=1cm, font=\tiny, align=center]
      (s1) at (10.6, -0.7) {$\mathcal{S}_1$};
\node[draw=blue!45, fill=blue!4, rounded corners=2pt,
      minimum height=0.5cm, minimum width=1cm, font=\tiny, align=center]
      (s2) at (10.6, -1.8) {$\mathcal{S}_2$};
\node[draw=blue!45, fill=blue!4, rounded corners=2pt,
      minimum height=0.5cm, minimum width=1cm, font=\tiny, align=center]
      (sk) at (10.6, -2.9) {$\mathcal{S}_K$};

\node[font=\small, text=black!40] at (10.6, -2.25) {$\vdots$};

\draw[arr, color=blue!40] (mmd1.east) -- (s1.west);
\draw[arr, color=blue!40] (mmd2.east) -- (s2.west);
\draw[arr, color=blue!40] (mmdk.east) -- (sk.west);

\node[font=\tiny, text=blue!45, above right=0.02cm and 0.02cm of s1] {$n \!\ll\! N$};
\node[pfnbox, minimum width=1.6cm, minimum height=0.55cm] 
      (pfn1) at (13.2, -0.7) {{\tiny \textbf{PFN}$({\mathcal{S}_1, C_1})$}};
\node[pfnbox, minimum width=1.6cm, minimum height=0.55cm] 
      (pfn2) at (13.2, -1.8) {{\tiny \textbf{PFN}$({\mathcal{S}_2, C_2})$}};
\node[pfnbox, minimum width=1.6cm, minimum height=0.55cm] 
      (pfnk) at (13.2, -2.9) {{\tiny \textbf{PFN}$({\mathcal{S}_K, C_K})$}};

\node[font=\small, text=black!40] at (13.2, -2.25) {$\vdots$};

\node[font=\tiny, text=green!45!black, above=0.08cm of pfn1] {\textbf{Stage 3}};

\draw[arr, color=blue!40] (s1.east) -- (pfn1.west);
\draw[arr, color=blue!40] (s2.east) -- (pfn2.west);
\draw[arr, color=blue!40] (sk.east) -- (pfnk.west);

\node[outbox] (out1) at (15.6, -0.7) {{\tiny $\hat{\bm{y}}_{C_1}$}};
\node[outbox] (out2) at (15.6, -1.8) {{\tiny $\hat{\bm{y}}_{C_2}$}};
\node[outbox] (outk) at (15.6, -2.9) {{\tiny $\hat{\bm{y}}_{C_K}$}};

\node[font=\small, text=black!40] at (15.6, -2.25) {$\vdots$};

\draw[arr] (pfn1.east) -- (out1.west);
\draw[arr] (pfn2.east) -- (out2.west);
\draw[arr] (pfnk.east) -- (outk.west);

\node[draw=cyan!30, rounded corners=2pt, fill=cyan!3,
      font=\tiny, align=center, inner sep=3pt] at (1.5, -3.0) {%
  $\displaystyle\mathcal{S}_k^* = \arg\min_{|\mathcal{S}|=n}
  \mathrm{MMD}^2\!\big(\hat{P}_{C_k},\, \hat{P}_{\mathcal{S}}\big)$
};

\end{tikzpicture}%
}
\caption{%
\textbf{Overview of CRUMB.}
\textbf{Stage~1:} Test queries are partitioned into $K$ clusters via $k$-means.
\textbf{Stage~2:} For each cluster $C_k$, a training subset $\mathcal{S}_k$ of size $n \ll N$ is selected by greedy MMD minimisation (kernel herding), drawing from the full training pool (dashed blue arrows).
\textbf{Stage~3:} The PFN runs $K$ independent forward passes, each with a small, geometrically relevant context.
The total attention cost reduces from $T \!\cdot\! N$ to $T \!\cdot\! n$, enabling inference on datasets with $N > 50{,}000$ training points.
}
\label{fig:CRUMB_overview}
\end{figure}
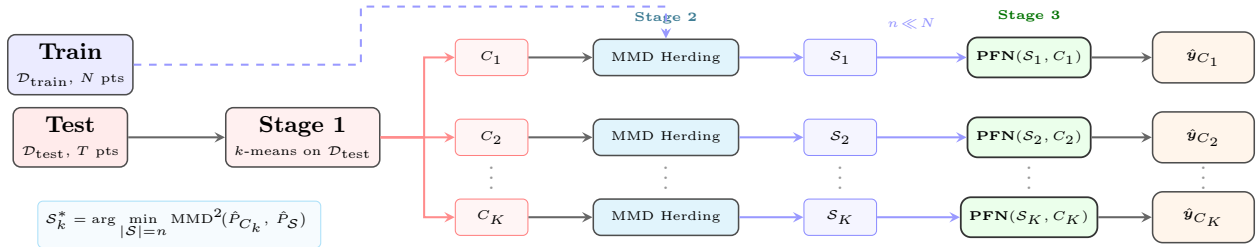

We propose \textbf{CRUMB} (\textbf{C}lustered \textbf{R}etrieval \textbf{U}sing \textbf{M}inimised-MMD \textbf{B}atching), a method that resolves this tension between context quality and batching efficiency.  The key idea is to cluster the \emph{test} queries, and to select a training context for each test cluster by minimising the maximum mean discrepancy
(MMD) between the test cluster and the selected train subset. This aligns the training context to the test query points that we wish to evaluate, while simultaneously allowing efficient PFN inference on batches of test queries at a time. An overview of CRUMB is given in Figure~\ref{fig:CRUMB_overview}.

Our contributions are as follows:

\begin{itemize}
    \item \textbf{MMD-based context retrieval.}  We propose greedy MMD
    minimisation (kernel herding) as the mechanism for selecting training
    subsets, and demonstrate that MMD-based selection
    outperforms centroid-nearest-neighbour and Voronoi-uniform retrieval 
    within the same clustering framework. 
    We also show that this MMD-minimised context retrieval provides 
    additional resilience to covariate drift in the test data compared 
    to other methods.

    \item \textbf{Batched context selection via test-side clustering.}  We
    show that clustering test queries and sharing a single training context
    per cluster enables efficient batching while preserving query-relevant
    context.

    \item \textbf{Strong results on TabArena.}  On the 51-dataset
    TabArena benchmark~\cite{erickson2025tabarenalivingbenchmarkmachine}, 
    evaluated across three PFN architectures, we find that CRUMB is close in performance to per-query $k$NN while requiring a fixed $K$ number of forward passes, instead of a forward pass for every testing point.
    CRUMB significantly outperforms both the Mixture of In-context Prompters (MICP) technique and uniform subsampling at identical context budgets. We also show that under covariate shift, CRUMB's advantage over MICP grows from $+4.9 \%$, to $+17.1\%$\, as drift intensifies from no drift, to completely out of sample covariate drift.
\end{itemize}

\section{Related Work}\label{sec:related}

\paragraph{Efficient Prior-Fitted Networks.} The PFN paradigm was introduced by \cite{hollmann2023tabpfntransformersolvessmall}, who showed that a transformer trained on synthetic datasets sampled from a prior can perform in-context learning on real tabular data. TabPFNv2 \cite{sergazinov2025chunkedtabpfnexacttrainingfree} and TabPFNv2.5 \cite{grinsztajn2026tabpfn25advancingstateart} scale this idea to larger and more diverse priors, achieving strong results on small-to-medium datasets \cite{mcelfresh2023neural}. The TabICL line of work \cite{qu2025tabicltabularfoundationmodel, qu2026tabiclv2betterfasterscalable} explores alternative architectures and training procedures within the same paradigm. All of these models share the fundamental scaling limitation that motivates our work: their inference-time cost scales quadratically with the number of training samples included in the context. Several works introduce architectural changes to improve speed, performance, and token efficiency, including TabFlex \cite{zeng2025tabflexscalingtabularlearning}, MITRA \cite{zhang2025mitra}, and related TabPFN variants \cite{kolberg2025tabpfn}. Chunked TabPFN \cite{sergazinov2025chunkedtabpfnexacttrainingfree} introduces a tiled block-attention strategy that partitions the attention tensor into chunks to compute full attention incrementally, yielding attention-computation speedups. Moreover, some methods attempt to reduce the effective dimensionality of the dataset \cite{feuer2023scaling}, providing directions orthogonal to inference-time context subselection. Drift-resilient TabPFN variants have been proposed to improve robustness to temporal distribution shifts \cite{helli2024drift}. These models can handle covariate and concept shift, but primarily by modifying the PFN pretraining prior to incorporate time-varying data-generating mechanisms, rather than via inference-time context selection.

\paragraph{Context selection for in-context learning.} Prior work on PFNs shows that $k$NN-based context selection is an effective choice \cite{hollmann2023tabpfntransformersolvessmall, thomas2024retrievalfinetuningincontext}. However, a key drawback of $k$NN-based approaches is that test queries cannot be efficiently batched, since each query may induce a distinct retrieved context. MixturePFN \cite{xu2024mixtureincontextprompterstabular} proposes Mixture of In-Context Prompters (MICP), which routes nearby test points to a shared local training context, enabling efficient batching while retaining locality. Other closely related work on context subselection includes \cite{ma2024context, lu2026ulead}; however, these methods do not use MMD-based heuristics. 

\section{Background} \label{sec:background}

We consider a standard tabular supervised learning setting. Let $\mathcal{D}_{\text{train}} = \{(\bm{x}_i, y_i)\}_{i=1}^{N}$ denote the labelled training set with $\bm{x}_i \in \mathbb{R}^d$ and $y_i \in \{1, \dots, C\}$, for classification and $y_i \in \mathbb{R}$ for regression; let $\mathcal{D}_{\text{test}} = \{\bm{x}_j^*\}_{j=1}^{T}$ denote the test queries. A PFN takes the full training set as context and produces predictions for all test queries in a single forward pass $\mathbf{\hat{y}}= \text{PFN}(\mathcal{D}_{\text{train}}, \mathcal{D}_{\text{test}}  )$, without requiring any dataset specific training on $\mathcal{D}_{\text{train}}$. Instead PFN models have been previously pre-trained over a large range of synthetic datasets to approximate the posterior predictive distribution, which allows them to perform in-context learning in a single forward pass \cite{hollmann2023original2023}.
However, the cost of this forward pass is often quadratic in the number of training samples, due to self-attention between training samples \cite{hollmann2023original2023, qu2025tabicltabularfoundationmodel}, which becomes prohibitive when $N$ is large.

\paragraph{Maximum mean discrepancy.} The maximum mean discrepancy (MMD) is a kernel-based distance between probability distributions \cite{gretton2012}. Given two distributions $P$ and $Q$ and a reproducing kernel Hilbert space (RKHS) with kernel $\kappa$, the squared MMD is defined as

\begin{equation} \text{MMD}^2(P, Q) = \mathbb{E}_{\bm{x}, \bm{x}' \sim P}[\kappa(\bm{x}, \bm{x}')] - 2\mathbb{E}_{\bm{x} \sim P, \bm{z} \sim Q}[\kappa(\bm{x}, \bm{z})] + \mathbb{E}_{\bm{z}, \bm{z}' \sim Q}[\kappa(\bm{z}, \bm{z}')]. \end{equation}
Intuitively, $\text{MMD}^2(P, Q) = 0$ if and only if $P = Q$ (for characteristic kernels such as the Gaussian RBF), making it a natural criterion for measuring how well a selected training subset represents a target distribution. In our setting, $P$ is the empirical distribution over a test cluster $C_k$ and $Q$ is the empirical distribution over a candidate training subset $\mathcal{S}_k \subset \mathcal{D}_{\text{train}}$. 

\paragraph{Covariate Drift.} The standard assumption is that training and test inputs are drawn from the same marginal distribution $P(\bm{x})$. Covariate drift refers to the setting where this assumption is violated: the test inputs are drawn from a shifted distribution $P_{\text{test}}(\bm{x}) \neq P_{\text{train}}(\bm{x})$, while the labelling mechanism $P(y|\bm{x})$ remains unchanged \cite{Shimodaira2000ImprovingPI, Yarabolu_2024}. Under such a shift, a training context sampled uniformly from $\mathcal{D}_{\text{train}}$ will reflect $P_{\text{train}}$ rather than $P_{\text{test}}$, potentially degrading predictions in regions of the feature space where the test queries concentrate. This motivates context selection methods that explicitly align the training context to the test distribution.

\paragraph{Mixture of In-Context Prompters (MICP).} The most closely related baseline to CRUMB is the Mixture of In-Context Prompters (MICP) component of MixturePFN~\cite{xu2024mixtureincontextprompterstabular}. MICP~\cite{xu2024mixtureincontextprompterstabular} clusters the training data, constructs a fixed support set per cluster, and routes test points to the nearest training centroid. The full MixturePFN system additionally includes a Context-Aware Fine-Tuning (CAPFN) stage that trains per-dataset adapter layers; we omit CAPFN throughout to isolate the effect of context selection and ensure a fair comparison with CRUMB, which likewise does not modify model weights.

\section{Methodology}\label{sec:methodology}

The method we propose has three stages: (i) cluster the test queries, (ii) retrieve a training subset for each cluster by minimising the maximal mean discrepancy (MMD) between train and test points, and (iii) run the PFN forward pass on each (cluster, subset) pair, where all points in a given pair can be batched together. We term this method \textbf{C}lustered \textbf{R}etrieval \textbf{U}sing \textbf{M}inimised-MMD \textbf{B}atching (CRUMB).
We also describe optional enhancements, such as an adaptive method of training context selection, whereby we select points until we see no significant improvement in MMD minimisation, saving time by avoiding large contexts with limited improvement in performance. 

\textbf{Stage 1: Clustering the Test Queries.}
We partition the test set into $K$ clusters $\{C_1, \dots, C_K\}$ using $k$-means on the input features $\{\bm{x}_j^*\}_{j=1}^{T}$ \cite{Lloyd1982LeastSQ, arthur2007}. The number of clusters $K$ is a hyperparameter that trades off between two extremes: $K = 1$ means we will consider one shared training context for all test points, while $K = T$ recovers per-query selection, where every test point gets a unique training context (i.e. as is the case in per-query $k$NN).
The clustering is performed on standardised features (zero mean, unit variance) using Euclidean distance. We use Lloyd's algorithm initialised with $k$-means++. 

\textbf{Stage 2: Training Subset Selection via MMD.}
For each cluster $C_k$, we want to find a training subset $\mathcal{S}_k \subset \mathcal{D}_{\text{train}}$ of size $n$ that is ``close'' to the test points in $C_k$ in a distributional sense. We use the maximum mean discrepancy (MMD) as our notion of closeness, supported by ablations in Appendix~\ref{sec:ablation_selection}. In our case minimising $\text{MMD}^2$ is equivalent to solving 
\begin{equation}
\label{eq:mmd_objective}
\mathcal{S}_k^* = \argmin_{\mathcal{S}_k \subset \mathcal{D}_{\text{train}},\, |\mathcal{S}_k| = n} \left[ \frac{1}{n^2} \sum_{\bm{x}_i, \bm{x}_{i'} \in \mathcal{S}_k} \kappa(\bm{x}_i, \bm{x}_{i'}) - \frac{2}{n |C_k|} \sum_{\bm{x}_i \in \mathcal{S}_k} \sum_{\bm{x}^*_j \in C_k} \kappa(\bm{x}_i, \bm{x}_j^*) \right],
\end{equation}

where the first term penalises redundancy among the selected points (encouraging diversity), while the second rewards proximity to the test cluster (encouraging relevance).

Minimising the MMD objective in Equation~\ref{eq:mmd_objective} is a combinatorial optimisation problem; we solve it greedily via kernel herding \cite{huszar2016optimallyweightedherdingbayesianquadrature, chen2012supersampleskernelherding}. Starting from $\mathcal{S}_k = \emptyset$ at $t=0$, then at each step $t$ we select the training point that minimises the following scoring function:
\begin{equation}
\label{eq:greedy_score}
\bm{x}_{i^*} = \argmin_{\bm{x}_i \in \mathcal{D}_{\text{train}} \setminus \mathcal{S}_k} \left[ -\frac{2}{|C_k|} \sum_{\bm{x}^*_j \in C_k} \kappa(\bm{x}_i, \bm{x}_j^*) + \frac{2}{t+1} \sum_{\bm{x}_{i'} \in \mathcal{S}_k} \kappa(\bm{x}_i, \bm{x}_{i'}) \right],
\end{equation}

where the first term rewards proximity to the test cluster (precomputed once in $O(N|C_k|)$ time) and the second penalises redundancy with already-selected points (updated incrementally as each point is added). In our application we use a Gaussian RBF kernel $\kappa(\bm{x}, \bm{z}) = \exp(-\|\bm{x} - \bm{z}\|^2 / 2\sigma^2)$ with bandwidth $\sigma$ set via the median heuristic, which avoids introducing additional hyperparameters. Since the Gaussian RBF kernel satisfies $\kappa(\bm{x}_i, \bm{x}_i) = 1$ for all $i$, the self-kernel term is constant across candidates and can be dropped from the selection criterion. 

The end result of MMD-minimisation is that for each batch of test points $C_k$ we have found a training subset $\mathcal{S}_k$ to use for all test points in the batch. Note that the subsets $\mathcal{S}_1, \dots, \mathcal{S}_K$ are not required to be disjoint. A training point that lies in a region of feature space relevant to multiple test clusters will naturally be selected for multiple subsets. This is desirable as it means that informative training points are reused where needed.

\textit{Accelerated greedy MMD minimisation.} In practice we speed up this greedy procedure by selecting the top $B$ scoring points at each iteration before recalculating the second term of Eq.~\eqref{eq:greedy_score}, instead of selecting strictly one point at a time. We also speed up kernel calculations via random Fourier features (RFF)\cite{NIPS2007_013a006f}. For large datasets we also implement early stopping in the procedure, in which we monitor the relative MMD 
improvement every $50$ selected points and halt when improvement falls 
below $10^{-4}$ for five consecutive checks. This allows the context size 
to adapt to the difficulty of each cluster: easy clusters terminate early 
with fewer points, reducing inference cost without a fixed budget. We use 
this variant in the large-dataset experiments reported in 
Figure~\ref{fig:exp_large_context_summary}. Further details and ablations on greedy MMD minimisation variants are provided in Appendix~\ref{apd:greedy-mmd-improve}.

\textbf{Stage 3: Running Batched PFN Inference.}
Given the $K$ pairs $(C_k, \mathcal{S}_k)$, we run the PFN forward pass $K$ times, each time providing $\mathcal{S}_k$ as the training context for the test points in $C_k$.
For each cluster $k$ we build a single PFN input consisting of the shared context $\mathcal{S}_k$ and a query tensor containing all points in $C_k$, so that the PFN produces $\lvert C_k\rvert$ predictions in one vectorized forward pass.
This enables efficient GPU utilisation because all queries in $C_k$ share the same context, whereas per-query retrieval (e.g.\ $k$NN) induces distinct contexts and therefore prevents straightforward batching.

\begin{figure}[t]
    \centering
\includegraphics[width=0.95\linewidth]{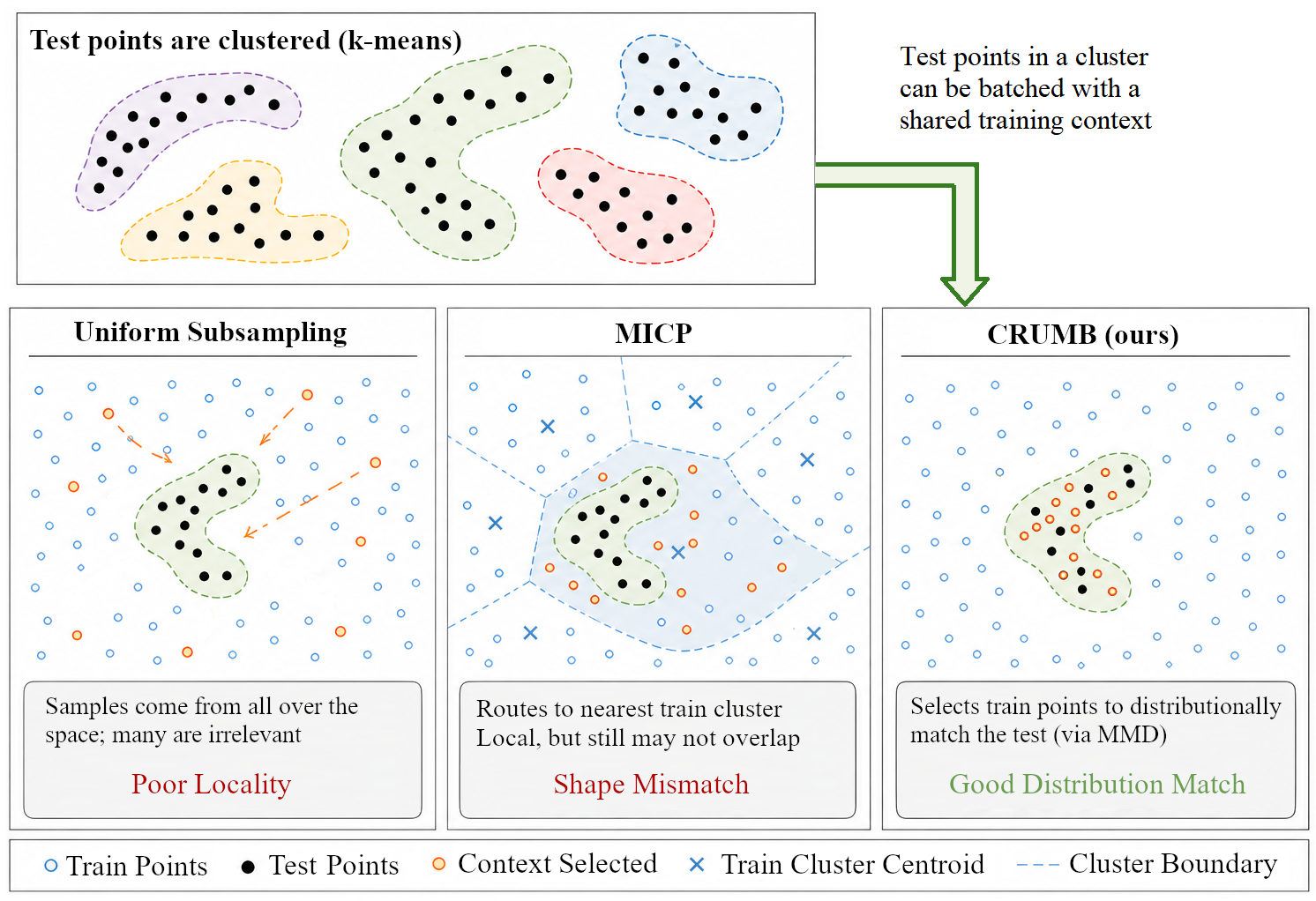}
    \caption{\textbf{Visual representation of the CRUMB preprocessing and context retrieval steps.} We also provide visualisation of alternative context retrievals, showing how uniform subsampling and MICP could in certain cases lead to a sub-optimal context selection for a given batch of test points.}
    \label{fig:MMD_retrieval_explanation}
\end{figure}

The full procedure is summarised in Algorithm~\ref{alg:crumb} and a visual representation of the preprocessing and context retrieval steps are shown in Figure~\ref{fig:MMD_retrieval_explanation}.

\begin{algorithm}[t]
\caption{CRUMB - Clustering Retrieval Using Minimised-MMD Batching}
\label{alg:crumb}
\begin{algorithmic}[1]
\REQUIRE Training set $\mathcal{D}_{\text{train}}$, test set $\mathcal{D}_{\text{test}}$, number of clusters $K$, subset size $n$, PFN model $f$
\STATE Standardise features of $\mathcal{D}_{\text{train}}$ and $\mathcal{D}_{\text{test}}$
\STATE $\{C_1, \dots, C_K\} \leftarrow k\text{-means}(\mathcal{D}_{\text{test}}, K)$
\FOR{$k = 1, \dots, K$}
    \STATE $\mathcal{S}_k \leftarrow \text{GreedyMMD}(\mathcal{D}_{\text{train}}, C_k, n)$ \hfill $\triangleright$ Eq.~\eqref{eq:greedy_score}
    \STATE $\hat{\bm{y}}_{C_k} \leftarrow f(\mathcal{S}_k, C_k)$ \hfill $\triangleright$ PFN forward pass
\ENDFOR
\RETURN $\{\hat{\bm{y}}_{C_k}\}_{k=1}^{K}$
\end{algorithmic}
\end{algorithm}

\section{Experiments}
\label{sec:experiments}

We evaluate CRUMB on the TabArena benchmark \cite{erickson2025tabarenalivingbenchmarkmachine}, which consists of 51 diverse tabular classification and regression datasets spanning various sizes, feature types, and difficulty levels. We test across three PFN architectures: TabPFNv2 \cite{hollmann2023tabpfntransformersolvessmall}, TabICLv1 \cite{qu2025tabicltabularfoundationmodel}, and TabICLv2 \cite{qu2026tabiclv2betterfasterscalable}. To investigate effects only relevant to very large datasets (beyond the sizes found in TabArena) we supplemented certain experiments with the Higgs particle physics dataset \cite{higgs_280}.

\paragraph{Baselines.} We restrict our investigation to context selection only (no fine-tuning) and report on the following context selection methods:
\textbf{Full context}: Is our baseline where the entire training set of size $N$ is used (no context selection used).
\textbf{Uniform subsampling}: $n$ training points are selected uniformly at random from the training set to be used as the context.
\textbf{$k$NN}: for each test point, the $n$ nearest neighbours in the training set are selected. Each test point receives a (potentially) unique training context and therefore they can not be batched.
\textbf{MICP}: the training data is clustered into $K' = \lceil \gamma N / n \rceil$ clusters via $k$-means, and a fixed support set of $n$ training points is constructed per cluster (by uniformly subsampling large clusters or expanding small ones via $k$NN from the centroid if the cluster does not include enough points) \cite{xu2024mixtureincontextprompterstabular}. 

\newpage
Each test point is routed to the nearest training-cluster centroid and batched with other test points assigned to the same cluster.

\paragraph{Metrics.} We report the ranking of different context selection techniques against each other, where the methods are ranked for a given PFN model and a dataset drawn with a fixed random seed. For classification tasks the rank is determined by the final accuracy; for regression tasks the rank is determined by RMSE. Wall-clock timings are reported only for the large-dataset experiments 
(Section~\ref{sec:adaptive_crumb}), where context selection provides a 
genuine computational saving. On the smaller TabArena datasets 
($N < 10{,}000$), fixed overhead of PFN model initialisations, forward passes, and preprocessing steps can make timing comparisons uninformative. Wall-clock inference times are reported using a single NVIDIA A100 GPU.

\subsection{Main Results}

We evaluate CRUMB against four baselines on all 51 TabArena datasets
using TabICL-v2: full-context inference, per-query
$k$-NN retrieval, MICP, and uniform subsampling. Every method except
full context operates within a fixed context budget $n = 0.1N$;
methods are ranked per (dataset, seed) pair and compared to CRUMB via
Wilcoxon signed-rank tests with Bonferroni correction.

\begin{table}[t]
\centering
\caption{Average rank of context selection methods on TabICL-v2 across 51 TabArena datasets (10 seeds each, 510 paired observations). Methods are ranked per (dataset, seed) group using accuracy for classification tasks and RMSE for regression tasks, then averaged across all 510 groups; lower rank is better. Significance vs.\ CRUMB is assessed via Wilcoxon signed-rank test with Bonferroni correction with adjusted score given by $p_{\text{adj}}$. Context budget $n = 0.1N$ for all methods except Full Context. $K{=}20$ test clusters for CRUMB and $K' = \lceil \gamma N / n \rceil$ train clusters for MICP, where $\gamma =1$ . In some cases some clusters may not have any test samples routed to them, in this case MICP only runs $K^*$ forward passes, where $K^*$ is the number of unique clusters which have test points routed to them. 
The scaling of the PFN inference attention with context size is listed to emphasise full context is prohibitive on large datasets due to $O(N^2)$ scaling, and to also highlight that per-query $k$-NN requires $T$ forward passes as opposed to CRUMB which requires $K$, which can be much smaller in practice. A full complexity analysis of methods is provided in Appendix~\ref{app:complexity_details}. }
\label{table:tabicl2main}
\begin{tabular}{lccl@{\hskip 6em}c@{\hskip 0.6em}c}
\toprule
& \multicolumn{3}{c}{\textit{Performance}} & \multicolumn{2}{c}{\textit{Computational Scaling}} \\
\cmidrule(lr){2-4} \cmidrule(lr){5-6}
\textbf{Method} & \textbf{Avg.\ Rank} & $p_{\text{adj}}$ & \textbf{Significant} & \textbf{Attn./pass} & \textbf{\# Passes} \\
\midrule
Full Context       & 2.202 $\pm$ 0.098 & $\sim\!10^{-19}$ & Yes & $O(N^2)$ & $1$  \\
per-query $k$-NN   & 2.782 $\pm$ 0.107          & $0.106$           & No     & $O(n^2)$ & $T$  \\
CRUMB              & 2.975 $\pm$ 0.096        & ---               & ---    & $O(n^2)$ & $K$  \\
MICP               & 3.304  $\pm$  0.105       & $\sim\!10^{-4}$   & Yes & $O(n^2)$ & $K^*$  \\
Uniform            & 3.737  $\pm$  0.105     & $\sim\!10^{-18}$  & Yes & $O(n^2)$ & $1$  \\
\bottomrule
\end{tabular}
\end{table}

\paragraph{Outcome.} In Table~\ref{table:tabicl2main} using the full context expectedly performs best, but it uses all $N$ training points and is infeasible for large $N$; it is included only as an upper bound. Among methods operating within a fixed context budget $n = 0.1N$, CRUMB is statistically close to per-query $k$-NN ($p_{\text{adj}} = 0.106$) despite requiring only $K{=}20$ forward passes rather than $T{=}200$, a $10\times$ reduction which is also confirmed experimentally ($35$\,s vs.\ $361$\,s average PFN inference time). CRUMB also significantly outperforms both MICP and uniform subsampling ($p_{\text{adj}} < 0.001$).

\subsection{Consistency Amongst PFN Models and Sampling Sizes}

We next vary three axes: (i) the PFN backbone (TabICL-v1, TabICL-v2,
TabPFN-v2) and (ii) the context budget $n = p N$, sweeping the sampling
proportions $p \in \{0.05, 0.1, 0.2, 0.5, 0.8\}$ (iii) max train size $N \in \{500, 1000, 2000\}$.  We restrict this
analysis to the 38 classification datasets in TabArena, since TabICL-v1 does not support regression.

\begin{figure}[h]
\centering
\includegraphics[width=\textwidth]{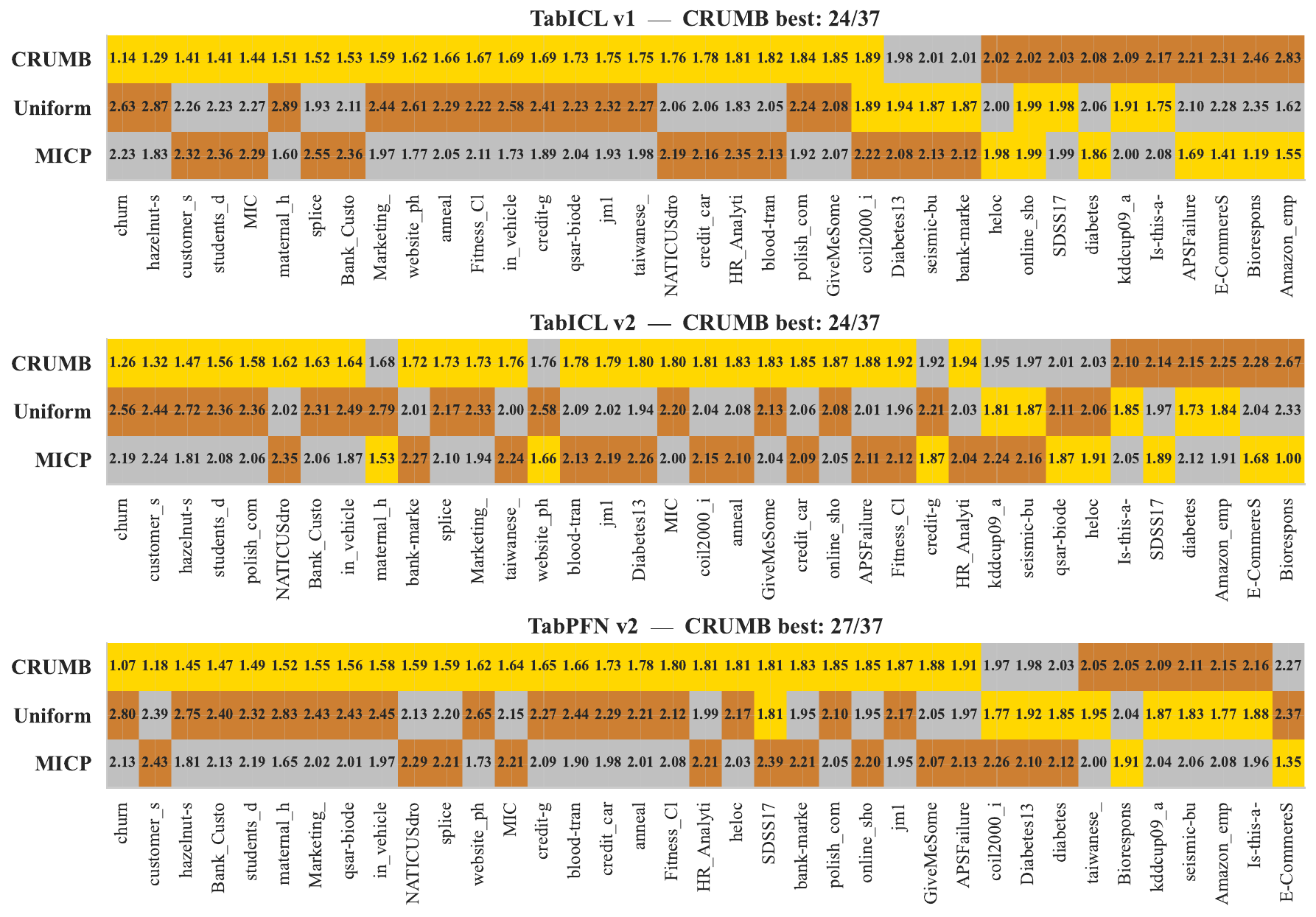}
\caption{\textbf{Average Rank Heatmap.}
For each dataset, we compute the mean accuracy of each context-selection method (CRUMB, Uniform, MICP) by averaging over all sampling proportions, all train sizes, and five random seeds. We report the mean rank for the methods for each dataset and PFN model combination. For each dataset and model colour indicates : Gold = best, Silver = Second, Bronze = Last.}
\label{fig:main_medal_heatmap}
\end{figure}

\begin{table}[t]
\centering
\caption{Average rank of context selection methods across the 38 TabArena classification datasets, 3 PFN models, 5 sampling proportions ($p \in \{0.05, 0.1, 0.2, 0.5, 0.8\}$), and 5 seeds. Methods are ranked per (dataset, model, proportion, seed) group using accuracy, then averaged; lower rank is better. Significance vs.\ CRUMB assessed via Wilcoxon signed-rank test with Bonferroni correction.}
\label{tab:medal_stats}
\begin{tabular}{lccl}
\toprule
\textbf{Method} & \textbf{Avg.\ Rank} & \textbf{$p_{\text{adj}}$} & \textbf{Significant} \\
\midrule
CRUMB    & $\mathbf{1.809} \pm 0.027$ & ---                        & ---   \\
MICP     & $2.022 \pm 0.023$          & $\sim\!10^{-8}$            & Yes \\
Uniform  & $2.169 \pm 0.026$          & $\sim\!10^{-9}$            & Yes \\
\bottomrule
\end{tabular}
\end{table}

\paragraph{Outcome.}
Figure~\ref{fig:main_medal_heatmap} and Table~\ref{tab:medal_stats}
confirm that CRUMB's advantage is not an artefact of a single model or
budget.  Across all three architectures, five sampling proportions and three train set sizes,
CRUMB achieves an average rank of $1.809 \pm 0.027$, significantly
outperforming both MICP ($2.022 \pm 0.023$ ) and uniform
subsampling ($2.169 \pm 0.026$ ).

\subsection{Large Datasets and Early Stopping}
\label{sec:adaptive_crumb}

We evaluate this method for large datasets where full-context PFN inference becomes computationally prohibitive due to quadratic attention cost. We run on the five largest TabArena datasets: Diabetes130US (57k), APSFailure (61k), SDSS17 (62k), customer\_satisfaction\_in\_airline (104k), and GiveMeSomeCredit (120k); capping the max size to 100k samples. We set the max context $n=0.1N$ and report results over 5 random seeds. In this experiment we enable early stopping of the greedy MMD minimisation procedure as detailed in Section \ref{sec:methodology}, when the MMD is no longer significantly decreasing.

\begin{figure}[h]
\centering
\includegraphics[width=\linewidth]{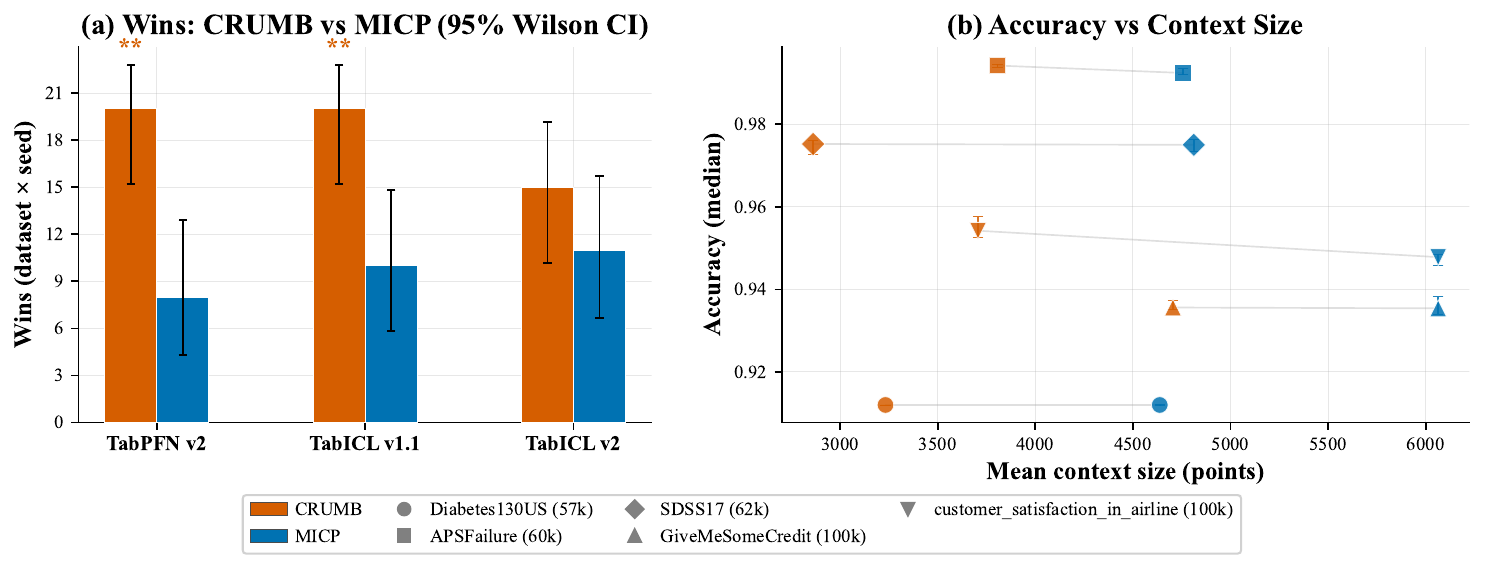}
\caption{\textbf{Large-dataset experiments (5 datasets, 5 seeds).}
\textbf{(a)}~Number of wins (dataset $\times$ seed) per context 
selection method, grouped by PFN model.
\textbf{(b)}~Accuracy (Median) vs.\ Mean context size for CRUMB and 
MICP across PFN models (marker shape distinguishes models); 
connecting lines link the two methods on the same dataset. ** indicates $p<0.01$. }
\label{fig:exp_large_context_summary}
\end{figure}

\paragraph{Outcome.}
Figure~\ref{fig:exp_large_context_summary}(a) shows the number of
wins (across dataset--seed pairs) for each method: CRUMB with early
stopping wins more often than MICP, significantly so for TabPFN-v2
and TabICL-v1 ($p<0.01$).
Figure~\ref{fig:exp_large_context_summary}(b) shows that the early
stopping criterion terminates context selection before the full
budget~$n$ is reached, yielding smaller contexts that perform
comparably to MICP when it is given a larger context size.

\subsection{Covariate Drift}\label{sec:covariate_shift}

We simulate controlled covariate drift on real datasets.
All features are standardised and projected onto the first principal component (PC1);
each sample receives a quantile rank $q \in [0,1]$ along PC1.
The training pool is fixed as the bottom half ($q < 0.5$; ${\sim}25{,}000$ points). This amounts to passing the data through a filter function, thus changing their distribution.
The test pool is passed through a similar filter but with an adjustable sliding window such that $q \in [\tau/2,\; 0.5 + \tau/2]$. This means that at $\tau{=}0$ the filters are the same for training and testing, hence this corresponds to no drift. In contrast, at $\tau{=}1$ the test dataset will be filtered to only include $q >0.5$, meaning that the test set will be entirely out-of-distribution as the train and test set will be in disjoint spaces (along the first PCA component axis). We adopted this method as a way to introduce parameterised covariate drift and out of distribution testing onto real datasets.

For each $\tau \in \{0.0, 0.1, \ldots, 1.0\}$ and random seed,
$T{=}200$ test points are sampled from the test pool.
Each method receives a context budget of $n = \min(0.1N,\; 10{,}000)$;
both CRUMB and MICP use $K{=}20$ clusters.
Results are averaged over $10$ seeds per $(\tau, \text{dataset}, \text{model})$ triple
across four datasets with $N \geq 50{,}000$:
two classification tasks from TabArena (Anneal, Amazon Employee Access),
one regression task (Airfoil Self Noise),
and the Higgs particle physics dataset \cite{higgs_280}.

\begin{figure}[t]
    \centering
    \begin{subfigure}[b]{0.48\textwidth}
        \centering
        \includegraphics[width=\textwidth]{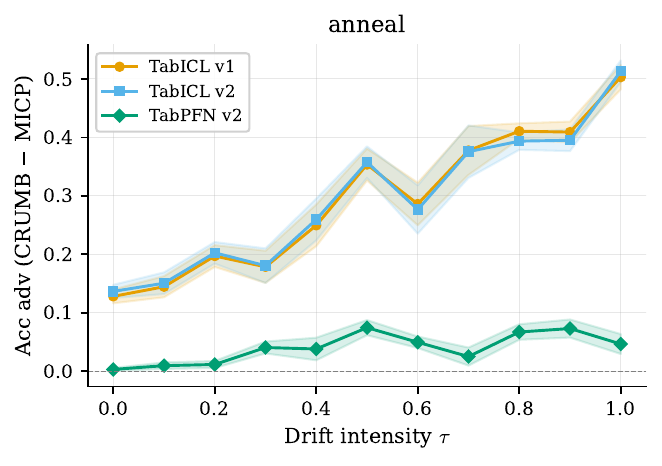}
        \label{fig:sub1}
    \end{subfigure}
    \hfill
    \begin{subfigure}[b]{0.48\textwidth}
        \centering
        \includegraphics[width=\textwidth]{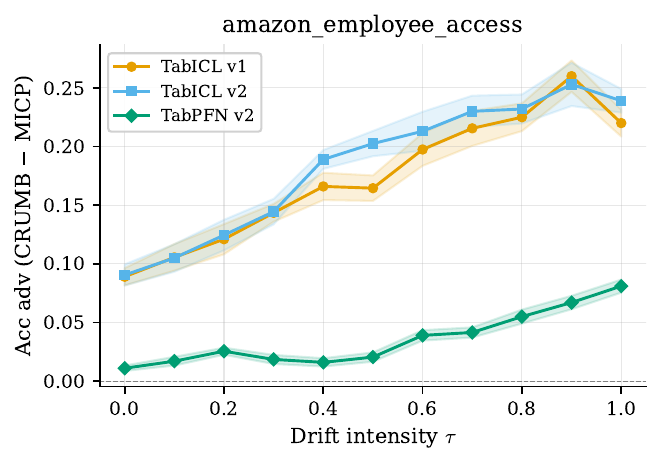}
        \label{fig:sub2}
    \end{subfigure}

    \vspace{0.01em}

    \begin{subfigure}[b]{0.48\textwidth}
        \centering
        \includegraphics[width=\textwidth]{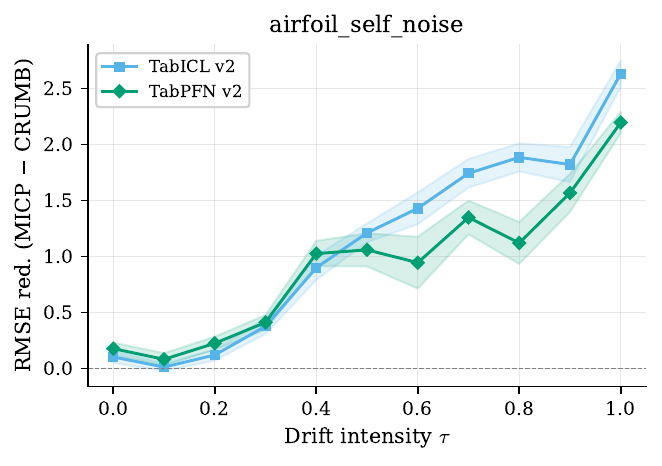}
        \label{fig:sub3}
    \end{subfigure}
    \hfill
    \begin{subfigure}[b]{0.48\textwidth}
        \centering
        \includegraphics[width=\textwidth]{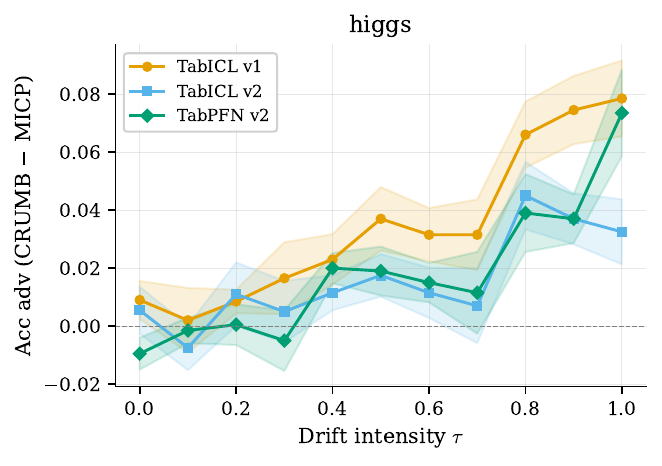}
        \label{fig:sub4}
    \end{subfigure}

    \caption{\textbf{CRUMB advantage under controlled covariate drift.} Each panel shows accuracy (or RMSE for Airfoil) versus drift intensity $\tau$ for CRUMB and MICP across PFN models. CRUMB's advantage widens as $\tau$ increases, demonstrating robustness to distribution shift. TabICLv1 is omitted from the regression panel as it does not support regression.}
    \label{fig:covariate_drift}
\end{figure}

\paragraph{Outcome.}
Table~\ref{tab:exp8_drift_clf} and Figure~\ref{fig:covariate_drift}
show that CRUMB's advantage over MICP grows monotonically with drift intensity.
At $\tau{=}0$, CRUMB leads by $+4.9$\%;
at $\tau{=}1$ the gap widens to $+17.1$\%,
as CRUMB degrades gracefully
($0.794$ to $0.699$; $-9.5$\%)
while MICP collapses
($0.746$ to $0.528$; $-21.8$\%).
The mechanism is straightforward:
MICP clusters the \emph{training} data, so its partitions become
misaligned with the shifting test distribution;
CRUMB clusters on the test side and adapts automatically,
providing a natural robustness mechanism against covariate drift.

\begin{table}[H]
\centering
\caption{Classification accuracy under covariate drift, aggregated across
3 datasets, 3 PFN models, and 10 seeds.  Values are mean $\pm$ SEM.}
\label{tab:exp8_drift_clf}
\begin{tabular}{@{}c ccc@{}}
\toprule
$\tau$ & CRUMB & MICP & $\Delta_{\text{CRUMB\textendash MICP}}$ \\
\midrule
0   & $\mathbf{0.794} \pm 0.029$ & $0.746 \pm 0.033$ & $+0.049 \pm 0.006$ \\
0.5 & $\mathbf{0.788} \pm 0.019$ & $0.663 \pm 0.031$ & $+0.125 \pm 0.015$ \\
1.0 & $\mathbf{0.699} \pm 0.014$ & $0.528 \pm 0.031$ & $+0.171 \pm 0.021$ \\
\bottomrule
\end{tabular}
\end{table}

\section{Conclusion}
\label{sec:conclusion}

We introduced CRUMB, a training-free inference wrapper that makes
prior-fitted network inference practical on large tabular datasets by
clustering test queries, selecting distributionally matched training
contexts via greedy MMD minimisation, and running batched PFN forward
passes on the resulting reduced-context pairs.
Across 51 TabArena datasets and three PFN architectures, CRUMB
significantly outperforms both uniform subsampling and MICP at
identical context budgets.
The source of this advantage is the combination of test-side clustering
with distributional context retrieval: clustering creates focused,
localised query groups, and MMD herding ensures each group receives a
context that is distributionally aligned rather than merely
geometrically proximal.
This same test-adaptive design provides natural resilience to covariate
drift, a regime in which CRUMB's advantage over MICP grows from
$+4.9$\% to $+17.1$\% as drift intensifies. 

 \paragraph{Limitations} CRUMB requires the full set of test queries upfront in order to cluster
them and does not natively support the online setting where points
arrive one at a time.
We discuss a cached-centroid variant for streaming inference in
Appendix~\ref{apd:onlinecrumb}, but this effectively reduces to
MICP-style routing with a different preprocessing step, and the
distribution-matching advantage of CRUMB may be diminished.
The clustering stage relies on Euclidean $k$-means, which assumes
roughly isotropic cluster geometry and may be poorly suited to
high-dimensional or mixed-type feature spaces; the MMD objective is
subject to the same limitation, using an isotropic Gaussian RBF kernel
with a fixed median-heuristic bandwidth and no adaptation to
dataset-specific structure.
Although CRUMB's preprocessing scales linearly in $N$, the constant in
$O(Nd(T{+}Kn))$ can still be large for million-scale datasets.
More broadly, as a pure inference wrapper CRUMB inherits the ceiling of
the underlying PFN: if the model's pretraining priors are a poor match
for a given dataset, no context selection strategy can fully compensate.
In this work we focused solely on context selection without additional
fine-tuning; methods such as LoCalPFN~\cite{thomas2024retrievalfinetuningincontext}
and MixturePFN~\cite{xu2024mixtureincontextprompterstabular} apply
fine-tuning after $k$NN and MICP respectively. 
Investigating
whether similar fine-tuning combined with CRUMB yields further gains is
a natural direction for future work.

\section*{Disclaimer}

This paper was prepared for informational purposes by the Global Technology Applied Research center of JPMorgan Chase \& Co. This paper is not a merchandisable/sellable product of the Research Department of JPMorgan Chase \& Co. or its affiliates. Neither JPMorgan Chase \& Co. nor any of its affiliates makes any explicit or implied representation or warranty and none of them accept any liability in connection with this paper, including, without limitation, with respect to the completeness, accuracy, or reliability of the information contained herein and the potential legal, compliance, tax, or accounting effects thereof. This document is not intended as investment research or investment advice, or as a recommendation, offer, or solicitation for the purchase or sale of any security, financial instrument, financial product or service, or to be used in any way for evaluating the merits of participating in any transaction.



\bibliographystyle{unsrtnat}
\bibliography{ref}

\newpage
\appendix


\section{Experimental Details}\label{app:experimental_details}

\subsection{Models}\label{app:models}

We evaluate three PFN architectures, all loaded from publicly released checkpoints without modification:

\begin{itemize}
  \item \textbf{TabPFNv2.} 12-layer, 6-head transformer ($d{=}192$,
    $d_h{=}32$), maximum context window $N_{\max}{=}10{,}000$.
    Supports classification and regression.

  \item \textbf{TabICLv1.} 12-layer, 4-head transformer ($d{=}512$,
    $d_h{=}128$), $N_{\max}{=}4{,}096$.
    Classification only.

  \item \textbf{TabICLv2.} 12-layer, 8-head transformer ($d{=}512$,
    $d_h{=}64$), $N_{\max}{=}4{,}096$.
    Supports classification and regression.
\end{itemize}

\subsection{Default Hyperparameters}\label{app:crumb_params}

Table~\ref{tab:crumb_params} lists the default CRUMB
hyperparameters used throughout the paper.  Any deviations are
stated in the relevant experiment subsection.

\begin{table}[h]
\centering
\caption{Default CRUMB hyperparameters.}
\label{tab:crumb_params}
\small
\renewcommand{\arraystretch}{1.05}
\begin{tabular}{@{}lll@{}}
\toprule
\textbf{Parameter} & \textbf{Value} & \textbf{Description} \\
\midrule
$K$ (test clusters)    & 20              & Number of $k$-means clusters \\
$D_{\mathrm{rff}}$     & 64              & Random Fourier Feature dimension \\
$B$ (greedy batch)     & 50              & Points selected per greedy iteration \\
$\sigma$               & median heuristic & RBF kernel bandwidth \\
\bottomrule
\end{tabular}
\end{table}

\subsection{Other Experimental Details}\label{app:reproducibility}

All experiments are run on a single NVIDIA A100 (80\,GB) GPU with
64 CPU cores; clustering and MMD selection run on CPU, PFN forward
passes run on GPU. Random seeds control train/test splits, $k$-means
initialisation, RFF projections, and stochastic subsampling.
Each method is subject to a per-run wall-clock timeout
(300\,s for main experiments; 1\,800\,s for large-context
experiments); timed-out runs are excluded from analysis.

\section{Ablation Studies}\label{sec:ablation_clustering}

We ablate each design choice of CRUMB in isolation: the number of test-side clusters~$K$, the per-cluster context budget~$n$, and the within-cluster retrieval strategy.
All ablations sweep a single hyperparameter while holding the remainder at their default values ($K{=}20$, $n{=}0.1N$, MMD herding) and report the normalised score averaged across all TabArena datasets and 10 seeds.

\subsection{Number of clusters $K$.}
\begin{figure}[H]
    \centering
    \begin{subfigure}[b]{0.49\linewidth}
        \centering
        \includegraphics[width=\linewidth]{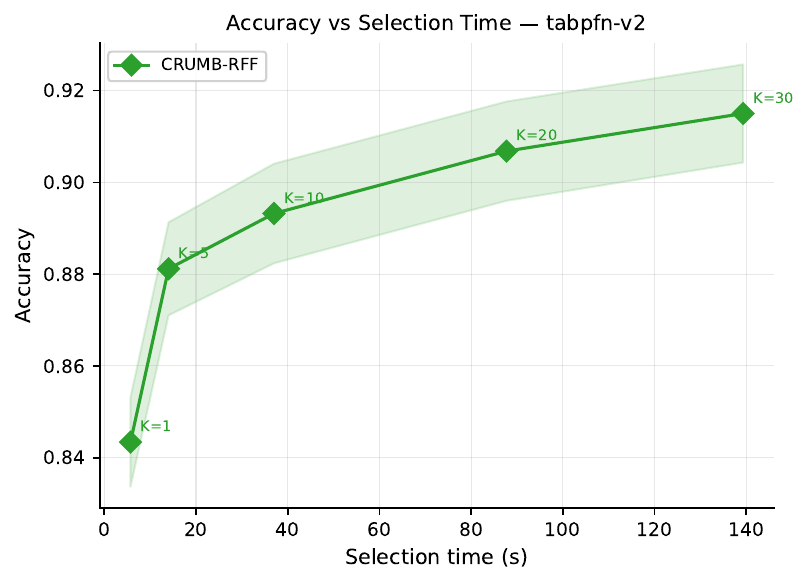}
        \caption{TabPFN\,v2}
    \end{subfigure}
    \hfill
    \begin{subfigure}[b]{0.49\linewidth}
        \centering
        \includegraphics[width=\linewidth]{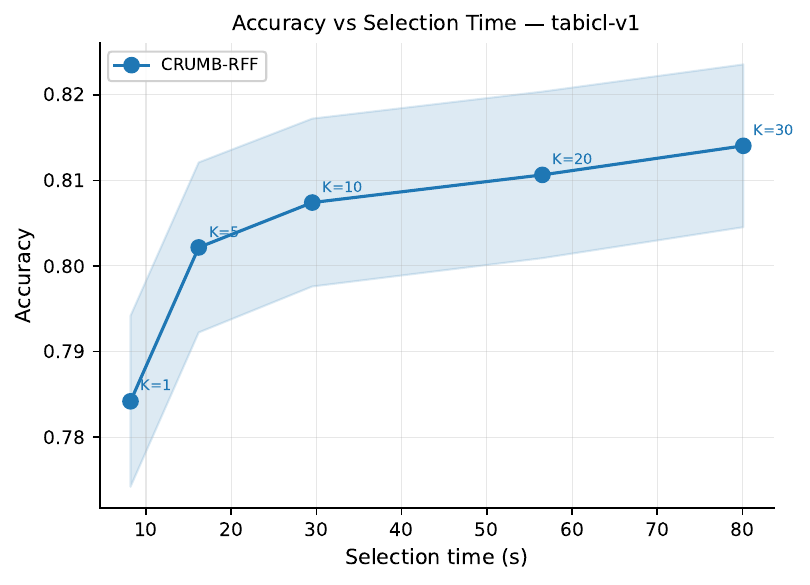}
        \caption{TabICL\,v1}
    \end{subfigure}
    
    \vspace{0.5em}
    
    \begin{subfigure}[b]{0.49\linewidth}
        \centering
        \includegraphics[width=\linewidth]{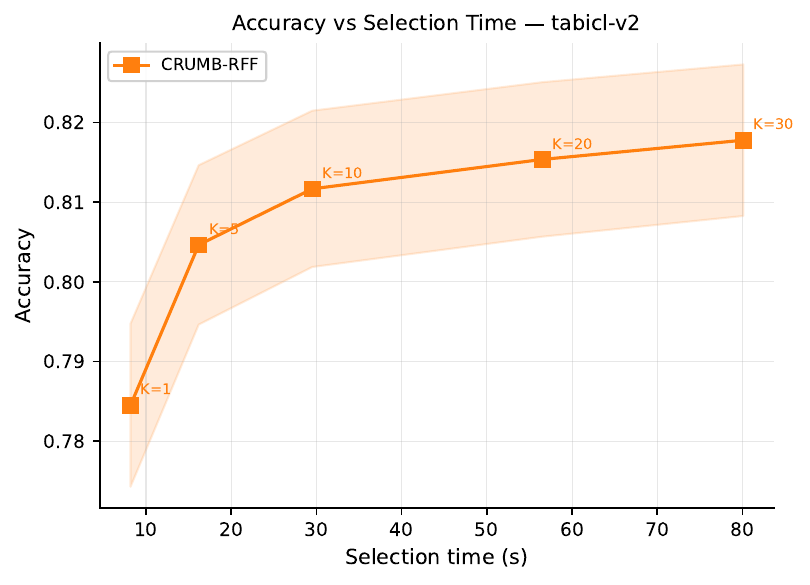}
        \caption{TabICL\,v2}
    \end{subfigure}
    \caption{Accuracy vs.\ selection time as the number of clusters~$K$ varies.  Each point is one value of $K$; error bars show $\pm1$ SEM over seeds. }
    \label{fig:ablation_num_cluster}
\end{figure}

Figure~\ref{fig:ablation_num_cluster} plots accuracy against selection time as $K$ varies in $\{1, 5, 10, 20, 30\}$ for fixed $n = 0.1N$.
Setting $K{=}1$ collapses CRUMB to a single global context: accuracy degrades because the selected subset cannot simultaneously represent all regions of the test distribution.
Performance improves as $K$ increases, but improvements start to become more marginal after $K=10$.

\subsection{Within-Cluster Retrieval Strategy}\label{sec:ablation_selection}

All variants of CRUMB share the same test-side clustering step via $k$-means \cite{Lloyd1982LeastSQ}; they differ only in how the $n$ training points are chosen for each cluster.
We compare three strategies:

\begin{enumerate}
    \item \textbf{MMD herding} (default).  Greedy kernel herding that iteratively selects the training point minimising the empirical MMD between the growing context set and the test cluster (Algorithm~\ref{alg:crumb}).

    \item \textbf{Centroid-NN.}  Select the $n$ training points nearest (in $\ell_2$) to the test-cluster centroid.
    This mirrors the routing mechanism of MICP, but applied to test-side rather than train-side clusters.

    \item \textbf{Voronoi-Uniform.}  The $K$ test-cluster centroids induce a Voronoi partition of the training set: each training point is assigned to its nearest centroid.
    For cluster $k$, we then uniformly subsample $n$ points from the Voronoi cell $\mathcal{V}_k = \{x_i \in \mathcal{D}_{\mathrm{train}} : k = \arg\min_{j} \lVert x_i - c_j \rVert\}$.
    This guarantees that the selected context lies in the same region of feature space as the test cluster, but without any distributional objective.
    We consider this as a benchmark to have a midpoint between the $k$NN's locality and the spread of uniform subsampling.
\end{enumerate}

All three strategies use the same $K{=}20$ clusters, context budget $n{=}500$, and training pool (capped at $N{=}10{,}000$), and are evaluated on the full TabArena collection with 10 seeds per dataset.

\begin{figure}[t]
    \centering
    \begin{subfigure}[b]{0.7\linewidth}
        \centering
        \includegraphics[width=\linewidth]{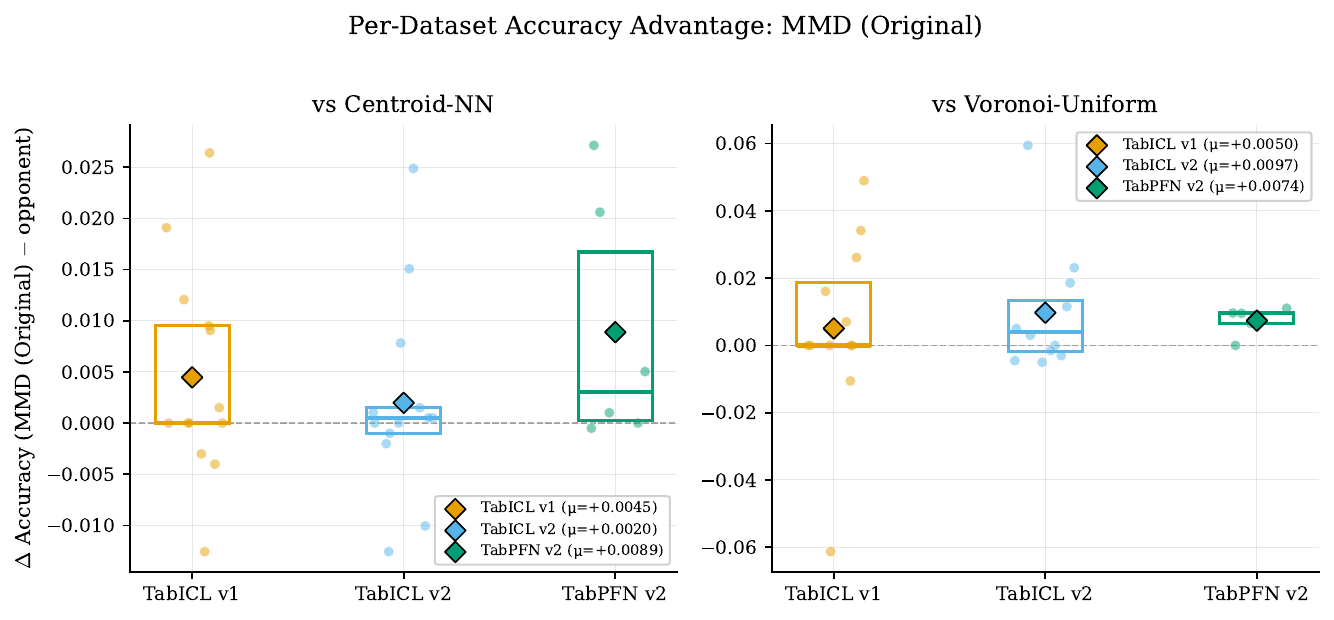}
        \caption{Accuracy advantage.}
        \label{fig:mmd_accuracy}
    \end{subfigure}
    \vfill
    \begin{subfigure}[b]{0.7\linewidth}
        \centering
        \includegraphics[width=\linewidth]{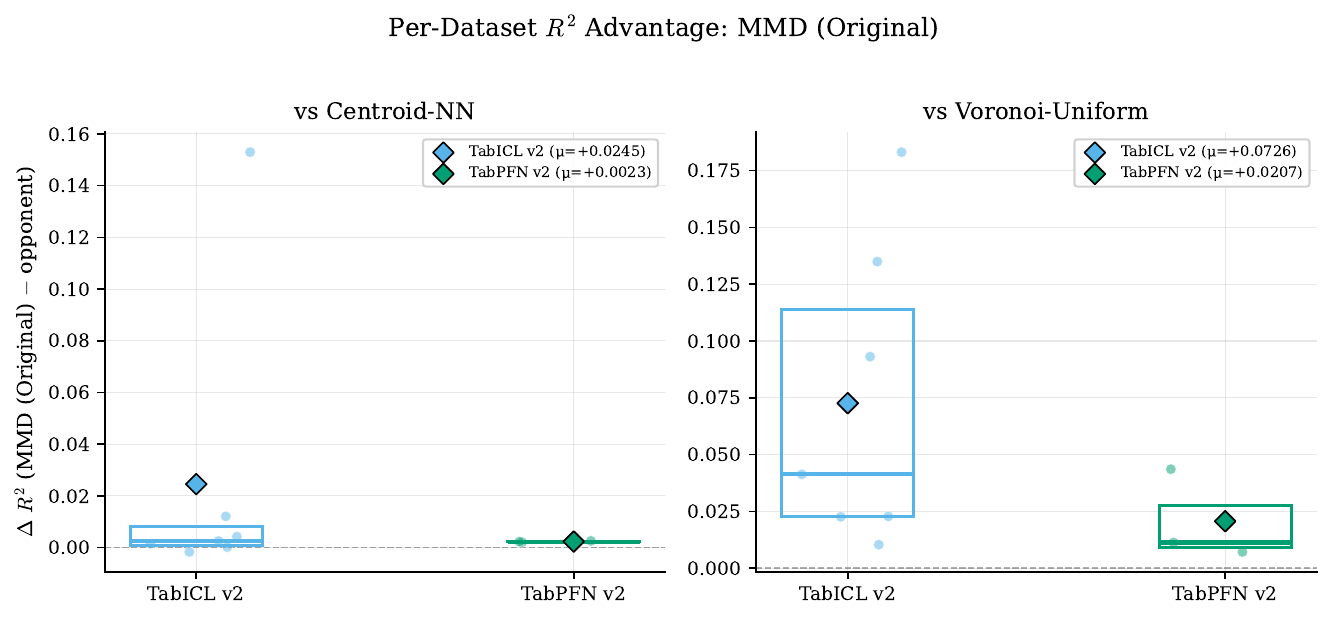}
        \caption{$R^2$ advantage.}
        \label{fig:mmd_r2}
    \end{subfigure}
    \caption{Per-dataset advantage of exact MMD herding over Centroid-NN and Voronoi-Uniform.  Each dot is one dataset's mean over 10 seeds; boxes show the IQR; diamonds mark the per-model mean.  Positive values favour MMD herding.}
    \label{fig:mmd_context_selection}
\end{figure}

\begin{figure}[t]
    \centering
    \begin{subfigure}[b]{0.7\linewidth}
        \centering
        \includegraphics[width=\linewidth]{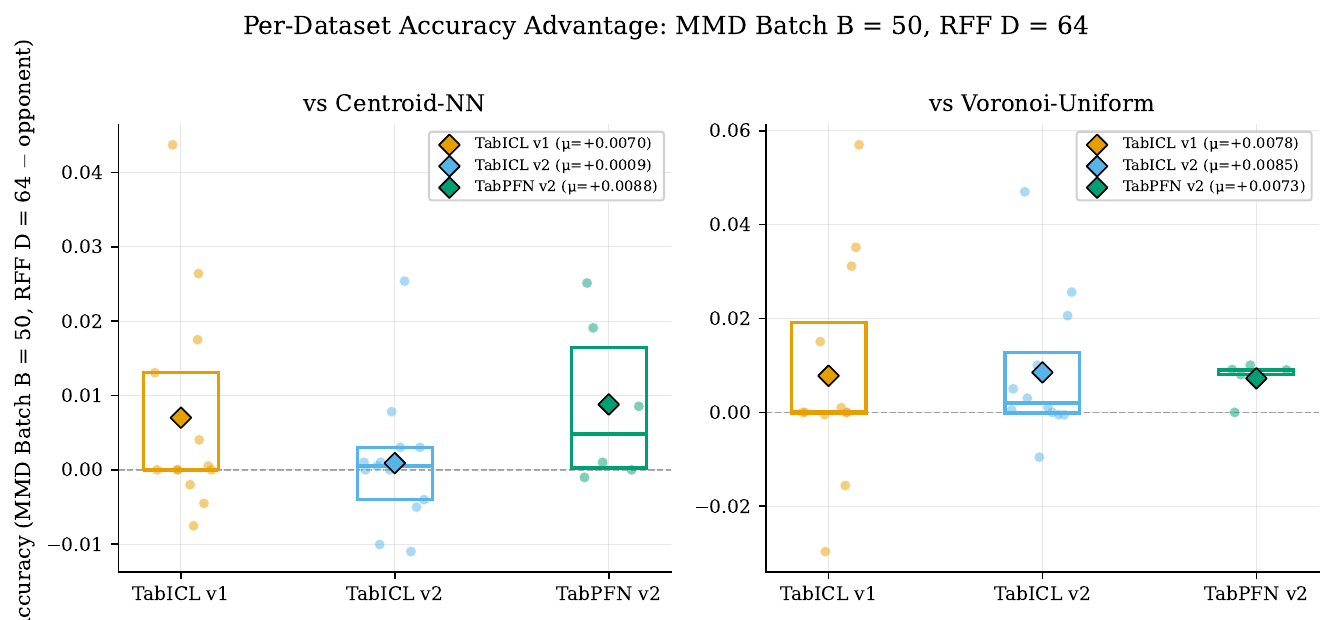}
        \caption{Accuracy advantage.}
        \label{fig:mmd_rff_accuracy}
    \end{subfigure}
    \vfill
    \begin{subfigure}[b]{0.7\linewidth}
        \centering
        \includegraphics[width=\linewidth]{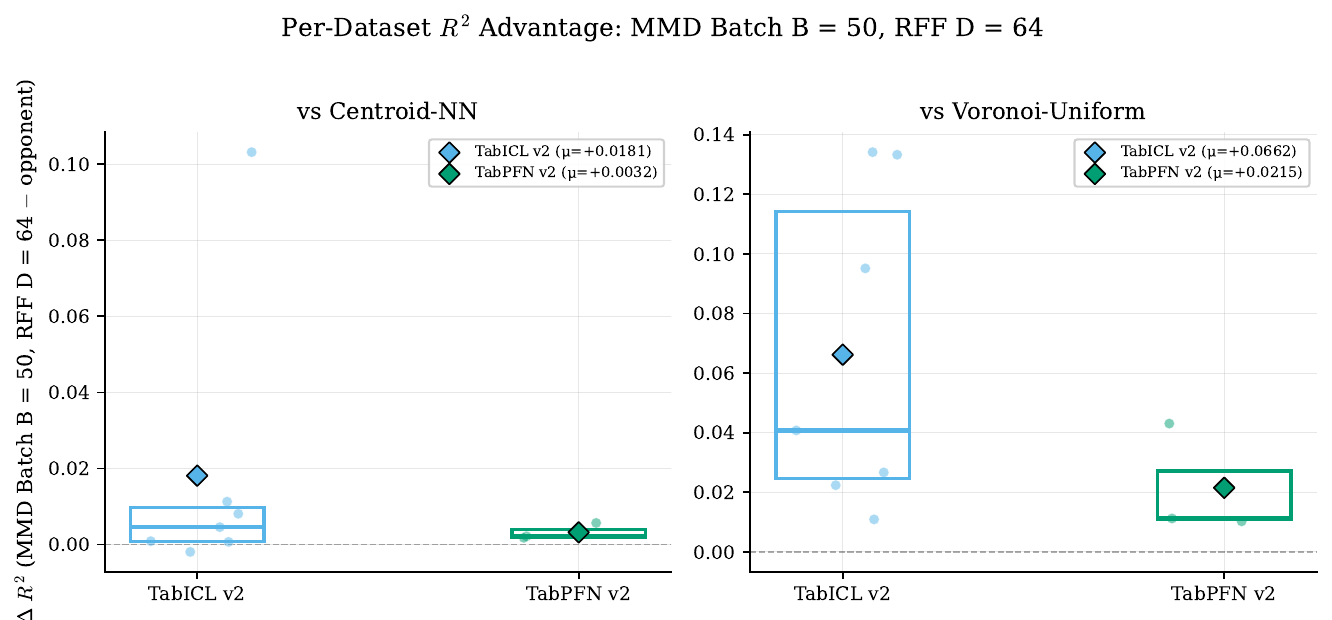}
        \caption{$R^2$ advantage.}
        \label{fig:mmd_rff_r2}
    \end{subfigure}
    \caption{Same comparison as Figure~\ref{fig:mmd_context_selection} but using the RFF-accelerated variant ($B{=}50$, $D{=}64$).  The advantage over both alternatives is preserved.}
    \label{fig:mmd_rff_context_selection}
\end{figure}

\paragraph{Results.}
Figures~\ref{fig:mmd_context_selection} and~\ref{fig:mmd_rff_context_selection} show the per-dataset accuracy and $R^2$ advantage of MMD herding over each alternative, displayed as strip plots with inter-quartile-range boxes per model.
MMD herding yields a positive mean advantage over both Centroid-NN and Voronoi-Uniform on every model, for both classification and regression tasks.

Centroid-NN tends to over-concentrate the context around the centroid, under-representing the tails of the test cluster.
Voronoi-Uniform selects points from the correct region but allocates budget uniformly rather than targeting distributional coverage, leading to redundancy in dense regions and gaps in sparse ones.
MMD herding avoids both failure modes by explicitly minimising the distributional mismatch between context and test cluster.

Figure~\ref{fig:mmd_rff_context_selection} confirms that the RFF-accelerated variant ($B{=}50$, $D{=}64$) preserves the advantage of exact MMD herding over both alternatives.

\subsection{Disentangling Clustering from Context Retrieval}
\label{ssec:cluster-vs-retrieval}

CRUMB comprises two pre-inference components: (i)~clustering the test points into
localised batches, and (ii)~retrieving a context subset for each batch
via greedy MMD minimisation.  A natural question is whether the
improvement stems from clustering, from context retrieval, or from
their interaction.  We isolate each factor with a $2\!\times\!2$
factorial design:

\begin{center}
\renewcommand{\arraystretch}{1.15}
\begin{tabular}{@{}lcc@{}}
\toprule
& \textbf{Uniform context} & \textbf{MMD context} \\
\midrule
\textbf{No clustering} & Case~A (baseline) & Case~C \\
\textbf{Clustering}     & Case~B            & Case~D (CRUMB) \\
\bottomrule
\end{tabular}
\end{center}

\noindent
\textbf{Case~A} is standard uniform subsampling: the full test set is
passed as a single batch with a uniformly drawn context.
\textbf{Case~B} clusters the test points but draws each cluster's
context uniformly from the entire training pool; since the context
distribution is identical across clusters, we expect this to be statistically
equivalent to Case~A.
\textbf{Case~C} skips clustering and applies MMD herding directly to
the full test set.  When the test set is large enough to approximate
the training distribution, the discrepancy is already small and we expect MMD
to yield little improvement over uniform sampling.
\textbf{Case~D (CRUMB)} is the full CRUMB pipeline: each test cluster defines
a localised target distribution that typically differs from the
training marginal, giving MMD herding the distributional gap it needs
to exploit.

We evaluate all four cases across the 51 TabArena datasets $\times$ 3
PFN models $\times$ 10 seeds $\times$ 3 context sizes
($n\!\in\!\{500, 1000, 2000\}$), with training pools capped at
$10{,}000$ points and $K\!=\!20$ clusters.

\begin{figure}[t]
    \centering
    \includegraphics[width=0.85\textwidth]{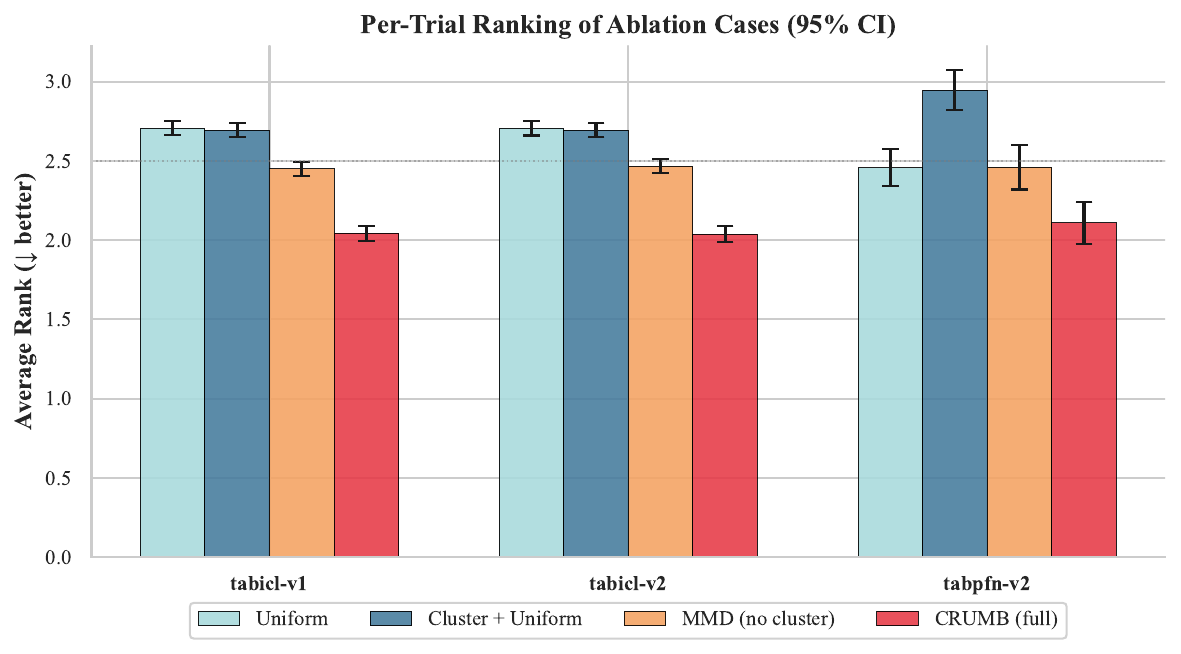}
    \caption{\textbf{Per-trial ranking of ablation cases (95\,\% CI).}
    Mean rank across all (dataset, model, seed) trials; error bars
    show 95\,\% confidence intervals.  The full CRUMB pipeline
    (Case~D: Cluster $+$ MMD) consistently achieves the best average rank,
    confirming that both components are important in improving performance.}
    \label{fig:exp11_rank}
\end{figure}

Figure~\ref{fig:exp11_rank} shows that Case~D achieves the best mean
rank.  Cases~A and~B perform similarly, as
expected, while Case~C offers only marginal improvement. The interaction between
clustering and retrieval is therefore the key mechanism: clustering
creates localised, distribution-shifted targets that MMD herding can
meaningfully serve.

\section{Online setting}\label{apd:onlinecrumb}

In this section, we highlight that in the online setting CRUMB can resolve into a variant of MICP. 
CRUMB, as described in the main text, operates in a test batch setting: it requires the full set of test queries upfront in order to cluster them. However, we show in Algorithm~\ref{alg:crumb_online} that the method adapts naturally to a streaming setting by caching the outputs of Stages 1 and 2. After running CRUMB on an initial batch of test points, we store the $K$ cluster centroids $\{\bm{\mu}_k\}$ and their associated training contexts $\{\mathcal{S}_k\}$. Subsequent test points are then routed to the nearest cached centroid and classified using the corresponding pre-computed context, with no further MMD computation required. 
Once enough points have been routed to a cluster to fill a batch, a single PFN forward pass is executed.

In this cached regime, CRUMB-Online \emph{reduces exactly to the MICP inference procedure}: both route each incoming test point to the nearest centroid and use a fixed, pre-computed training context. 
The algorithms are operationally identical at inference time. The only difference is in the \emph{initialisation}. 
Specifically, the algorithms differ in how the cached centroids and contexts are constructed. 
MICP's centroids are derived from clustering the \emph{training} data, and its contexts are constructed by subsampling or expanding training clusters. CRUMB-Online's centroids are derived from clustering recent \emph{test} queries, and its contexts are constructed via MMD minimisation against those test clusters. Once cached, both methods behave identically: route, look up context, run PFN.

This distinction becomes consequential under covariate shift. 
Because MICP's cluster structure is anchored to the training distribution, it remains fixed regardless of where future queries fall. In a similar manner the cached centroids of online CRUMB can become stale over time. A possible solution is that CRUMB-Online can periodically re-run the clustering and MMD selection steps on a sliding window of recent queries, refreshing both centroids and contexts to track distributional drift over time, an option unavailable to MICP with a fixed training set.

\begin{algorithm}[H]
\caption{CRUMB-Online: Streaming Inference with Cached Contexts}
\label{alg:crumb_online}
\begin{algorithmic}[1]
\REQUIRE Training set $\mathcal{D}_{\text{train}}$, initial test set $\mathcal{D}_{\text{test}}^{(0)}$, number of clusters $K$, subset size $n$, PFN model $f$
\STATE \textbf{// Phase 1: Initialisation (run CRUMB on initial test set)}
\STATE Standardise features of $\mathcal{D}_{\text{train}}$ and $\mathcal{D}_{\text{test}}^{(0)}$
\STATE $\{C_1, \dots, C_K\} \leftarrow k\text{-means}(\mathcal{D}_{\text{test}}^{(0)}, K)$
\FOR{$k = 1, \dots, K$}
    \STATE $\bm{\mu}_k \leftarrow \frac{1}{|C_k|} \sum_{\bm{x}^* \in C_k} \bm{x}^*$ \hfill $\triangleright$ Store cluster centroids
    \STATE $\mathcal{S}_k \leftarrow \text{GreedyMMD}(\mathcal{D}_{\text{train}}, C_k, n)$ \hfill $\triangleright$ Store training subsets
    \STATE $\hat{\bm{y}}_{C_k} \leftarrow f(\mathcal{S}_k, C_k)$ \hfill $\triangleright$ Predict initial test points
\ENDFOR
\STATE \textbf{Cache} $\leftarrow \{(\bm{\mu}_k, \mathcal{S}_k)\}_{k=1}^{K}$
\STATE
\STATE \textbf{// Phase 2: Streaming inference}
\STATE Initialise buffer $\mathcal{B} \leftarrow \emptyset$
\WHILE{new test points arrive}
    \STATE Receive new test point $\bm{x}^*_{\text{new}}$ (standardised)
    \STATE $k^* \leftarrow \arg\min_{k \in \{1,\dots,K\}} \|\bm{x}^*_{\text{new}} - \bm{\mu}_k\|^2$ \hfill $\triangleright$ Assign to nearest centroid
    \STATE $\mathcal{B}_{k^*} \leftarrow \mathcal{B}_{k^*} \cup \{\bm{x}^*_{\text{new}}\}$ \hfill $\triangleright$ Add to cluster buffer
    \IF{$|\mathcal{B}_{k^*}| \geq B_{\min}$ \textbf{or} timeout reached}
        \STATE $\hat{\bm{y}}_{\mathcal{B}_{k^*}} \leftarrow f(\mathcal{S}_{k^*}, \mathcal{B}_{k^*})$ \hfill $\triangleright$ Batched PFN forward pass
        \STATE \textbf{emit} $\hat{\bm{y}}_{\mathcal{B}_{k^*}}$
        \STATE $\mathcal{B}_{k^*} \leftarrow \emptyset$ \hfill $\triangleright$ Flush buffer
    \ENDIF
\ENDWHILE
\end{algorithmic}
\end{algorithm}

\section{Accelerations to Greedy MMD Minimisation Implementation}
\label{apd:greedy-mmd-improve}

In this section, we describe the optimizations we apply to the greedy
MMD herding in CRUMB to improve its runtime.
There are three main accelerations:
(1)~we approximate the Gaussian RBF kernel via \textit{Random Fourier
Features}~\cite{NIPS2007_013a006f}, replacing $O(d)$ distance
computations and exponentiations with $O(D)$ inner products in a
low-dimensional embedding space;
(2)~we use \textit{batched greedy selection}, choosing the top-$B$
scoring candidates per iteration before recomputing the redundancy
term, reducing the number of score-recomputation rounds from $n$ to
$\lceil n/B \rceil$;
(3)~we apply \textit{early stopping} to the herding procedure by
monitoring the relative decrease in $\widehat{\mathrm{MMD}}^2$ every
$\Delta$ selected points and halting when improvement falls below a
threshold $\varepsilon$ for $P$ consecutive checks, allowing simpler
clusters to terminate early with fewer context points and
correspondingly faster inference.

\textbf{Random Fourier Features.} The greedy MMD selection in Algorithm~1 requires repeated evaluation of the
Gaussian RBF kernel
$\kappa(x, x') = \exp\!\bigl(-\|x - x'\|^2 / 2\sigma^2\bigr)$
between candidate training points and the running selected set.
Na\"ively, each greedy round requires $\mathcal{O}(N)$ kernel evaluations
against the newly selected point, and the full selection of $n$ points
costs $\mathcal{O}(Nn)$ kernel evaluations (or $\mathcal{O}(N \lceil n/B
\rceil)$ with batch size~$B$).  Since each kernel evaluation involves
computing a pairwise distance and an exponentiation, this becomes the
computational bottleneck for large training pools.

We accelerate selection using \emph{Random Fourier Features}
(RFF)~\cite{NIPS2007_013a006f} to
approximate the RBF kernel via an explicit finite-dimensional feature map.
Specifically, we construct a randomised embedding
$z \colon \mathbb{R}^d \to \mathbb{R}^D$ such that
\begin{equation}
  \kappa(x, x') \approx z(x)^\top z(x'),
  \label{eq:rff-approx}
\end{equation}
where
\begin{equation}
  z(x) = \sqrt{\frac{2}{D}} \cos(\Omega^\top x + b),
  \quad
  \Omega \sim \mathcal{N}\!\bigl(0, \sigma^{-2} I_d\bigr),
  \quad
  b \sim \mathrm{Uniform}[0, 2\pi].
  \label{eq:rff-map}
\end{equation}
Here $\Omega \in \mathbb{R}^{d \times D}$ and $b \in \mathbb{R}^D$ are
sampled once and reused for all points, so the embedding is consistent
across training and test sets.

Let $Z \in \mathbb{R}^{N \times D}$ denote the RFF embeddings of the
$N$~training points, and let
$\bar{z}_{\mathcal{C}} = \frac{1}{|\mathcal{C}_k|}
\sum_{x^* \in \mathcal{C}_k} z(x^*)$
be the mean embedding of the test cluster~$\mathcal{C}_k$.
The two key quantities in the greedy objective become simple linear-algebraic operations in the RFF space:
\begin{itemize}
  \item \textbf{Cross-term} (affinity of each candidate to the cluster):
    $\frac{1}{|\mathcal{C}_k|} \sum_{j} \kappa(x_i, x_j^*)
    \approx z(x_i)^\top \bar{z}_{\mathcal{C}}$,
    computed as a single matrix--vector product $Z \, \bar{z}_{\mathcal{C}}
    \in \mathbb{R}^N$.
  \item \textbf{Running kernel sum} (redundancy with already-selected
    points): $\sum_{j \in \mathcal{S}} \kappa(x_i, x_j) \approx
    z(x_i)^\top \!\!\sum_{j \in \mathcal{S}} z(x_j)$.
    After selecting a batch of $B$~points, we update the running sum
    $\tilde{z}_{\mathcal{S}} \leftarrow \tilde{z}_{\mathcal{S}} +
    \sum_{j \in \text{batch}} z(x_j)$, a $D$-dimensional vector addition.
    The per-candidate scores are then $Z \, \tilde{z}_{\mathcal{S}}
    \in \mathbb{R}^N$.
\end{itemize}
Each greedy round thus costs $\mathcal{O}(ND)$ for the matrix--vector
products plus $\mathcal{O}(BD)$ for the running-sum update, compared to
$\mathcal{O}(NBd)$ distance computations and exponentiations in the
exact-kernel version.  Since $D \ll N$ in practice (we use $D = 64$),
this yields a substantial speedup.

\paragraph{Batched greedy selection.}
The standard kernel herding procedure selects one point per iteration,
recomputing all $N$ candidate scores after each addition.
We accelerate this by selecting the top-$B$ scoring candidates at each
iteration before recomputing the redundancy term
$\tilde{z}_{\mathcal{S}}$.  This reduces the number of score-recomputation
rounds from $n$ to $\lceil n/B \rceil$, at the cost of slightly greedier
(less sequential) decisions within each batch.  In practice we use $B{=}50$
(Table~\ref{tab:crumb_params}); the ablation in
Appendix~\ref{app:rff_ablation} confirms that this introduces negligible
accuracy loss compared to $B{=}1$.

\paragraph{Early Stopping to MMD herding.} 
In practice rather than selecting a fixed budget of $n$ context points,
the algorithm monitors the squared MMD value
$\widehat{\mathrm{MMD}}^{\,2}_t
= \bigl\|\bar{z}_{\mathcal{C}} - \tfrac{1}{|S_t|}
  \sum_{i \in S_t} z(x_i)\bigr\|^2$
every $\Delta{=}50$ selected points (after a warm-up of
$\max(100,\, n/10)$) and applies early stopping when the
relative improvement
$r_t = (\widehat{\mathrm{MMD}}^{\,2}_{t-\Delta}
       - \widehat{\mathrm{MMD}}^{\,2}_t)
      / \widehat{\mathrm{MMD}}^{\,2}_{t-\Delta}$
falls below $\varepsilon{=}10^{-4}$ for $P{=}5$ consecutive checks.
The final context size $|S_{t^*}| \leq n$ is thus determined
adaptively by the complexity of the local distribution:
simpler clusters converge faster, using fewer context points
and proportionally faster inference.
The overall cost is
$\mathcal{O}\!\bigl(\tfrac{t^*}{B}\, N D\bigr)$
where $t^* \leq n$ is the early-stop point.

\subsection{RFF and Batching Ablation}
\label{app:rff_ablation}

We compare 18 selection variants formed by crossing three axes:
kernel computation (exact vs.\ RFF with $D \in \{16, 64, 256\}$),
greedy batch size ($B \in \{1, 10, 50, 100\}$),
and two non-MMD baselines (Centroid-NN, Voronoi-Uniform).
All variants use $K{=}20$ clusters, context size $n{=}500$,
and are evaluated on the full TabArena collection
(38 classification + 13 regression datasets, 3 models, 10 seeds).

\paragraph{Key findings.}
The central result is that \emph{most MMD-based variants perform
comparably}, with only minor separation
between exact and some of the approximate configurations.
Non-MMD baselines (Centroid-NN, Voronoi-Uniform) rank noticeably
worse, confirming that the distributional objective itself, not the
precision of the kernel computation, drives the quality gains.

\begin{figure}[t]
\centering
\includegraphics[width=0.85\textwidth]{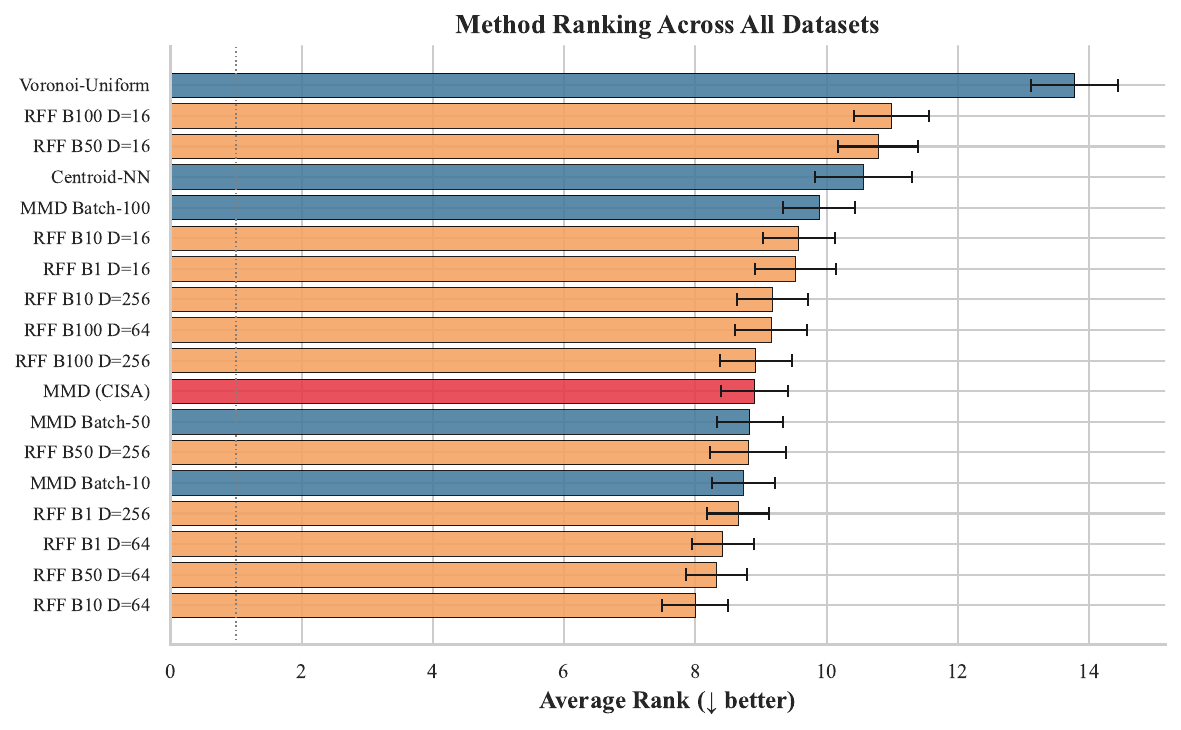}
\caption{\textbf{Average rank of all context-selection methods across (dataset, seed, model) trials.}
For each trial, methods are ranked by predictive performance (accuracy for classification, $R^2$ for regression) using average tie-breaking, with rank~1 being best.
Bars show the mean rank; error bars denote 95\% confidence intervals.}
\label{fig:rff_avg_rank}
\end{figure}

\begin{figure}[t]
\centering
\includegraphics[width=\textwidth]{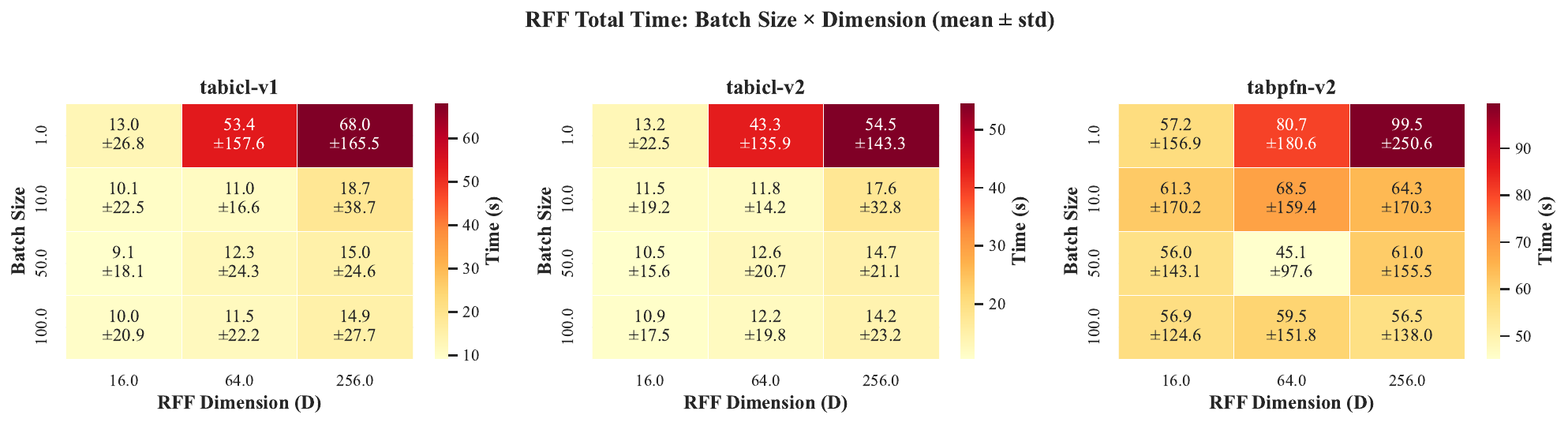}
\caption{\textbf{Wall-clock time of RFF approximations by batch size and feature dimension.}
Cells show mean total inference time $\pm$ standard deviation (in seconds) per model, aggregated over all datasets and seeds.}
\label{fig:rff_time_heatmap}
\end{figure}

\begin{figure}[t]
\centering
\includegraphics[width=0.85\textwidth]{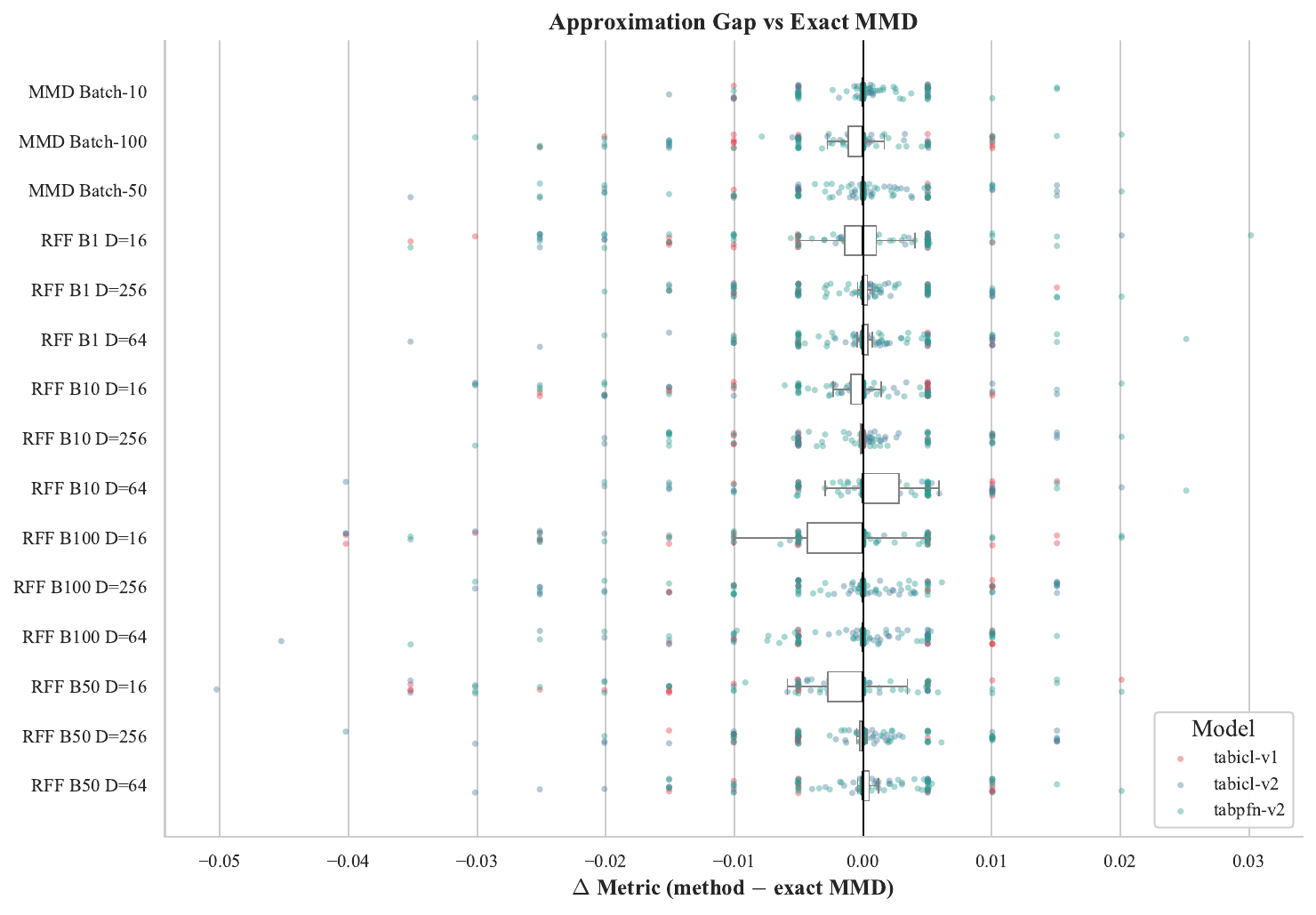}
\caption{Approximation gap ($\Delta = \text{method} - \text{exact MMD}$)
per trial. The performance profile (inset) reports the fraction of
trials within $\pm 1\%$, $\pm 2\%$, and $\pm 5\%$ of exact MMD.}
\label{fig:rff_delta}
\end{figure}

The practical implication is that RFF and batching can be used to
accelerate selection without sacrificing accuracy and with minimal hyperparameter tuning.
Figure~\ref{fig:rff_avg_rank} shows that mean accuracy ranking is reasonably
flat across the $(B, D)$ grid, while Figure~\ref{fig:rff_time_heatmap} shows
that larger batch sizes and small RFF dimension $D$ approximations reduce wall-clock.
The configuration $B{=}50$, $D{=}64$ is our recommended default:
it matches exact MMD accuracy to within ${\sim}0.2$\,pp while
running ${\sim}10{\times}$ faster.
Figure~\ref{fig:rff_delta} further corroborates this, showing that
the per-trial approximation gap is tightly concentrated around zero
for all RFF configurations.

\begin{figure}[t]
\centering
\includegraphics[width=\textwidth]{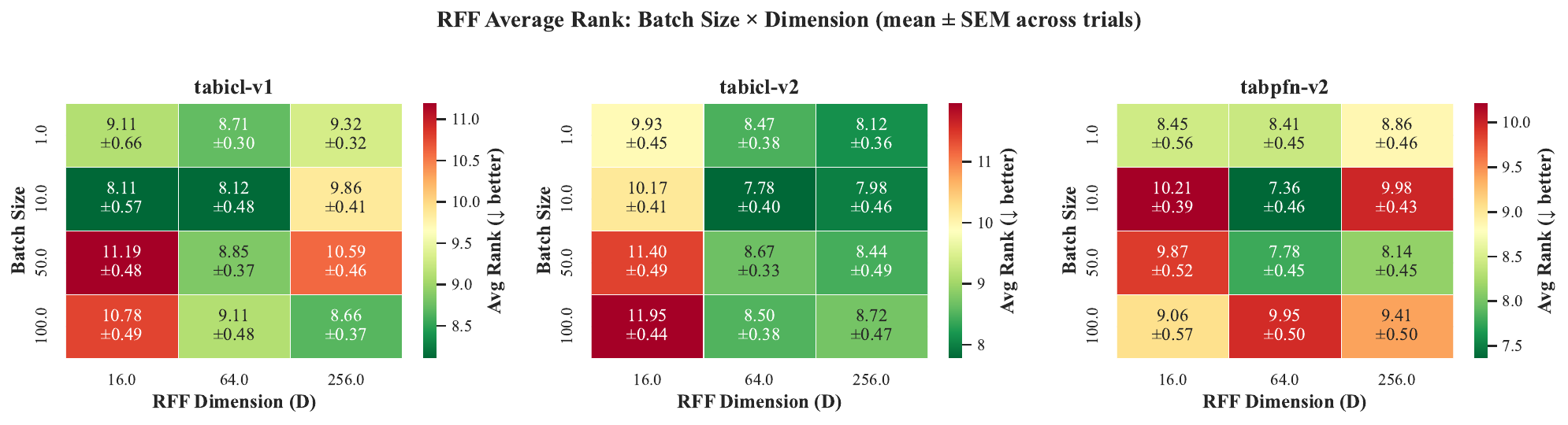}
\caption{\textbf{Average rank of RFF approximations by batch size and feature dimension.}
For each (dataset, seed, model) trial, all methods are ranked by predictive performance (rank~1 = best).
Cells show the mean rank $\pm$ SEM of each RFF configuration, aggregated over all trials.}
\label{fig:rff_rank_heatmap}
\end{figure}

\vspace{0.8em}

\clearpage

\section{Computational Complexity Analysis}\label{app:complexity_details}
In this section, we compare the computational complexity of the methods used in this paper. 
Table~\ref{tab:complexity_main} summarises the computational cost of each inference strategy along two axes: how the total cost scales with the training set size $N$ (for a fixed context budget $n$), and how many PFN forward passes are required.

\begin{table}[h!]
\centering
\caption{Computational complexity of inference strategies. We isolate scaling with training set size $N$ (for fixed $n, T, K$) and the number of PFN forward passes (which determines practical inference time). $^\dagger$MICP uses $K' = \lceil \gamma N/n \rceil$ prompters; with a cap on $K'$ the $N$-scaling reduces to $O(N)$. In some cases some clusters may not have any test samples routed to them, in this case MICP only runs $K^*$ forward passes, where $K^*$ is the number of unique clusters which have test points routed to them.}
\label{tab:complexity_main}
\small
\renewcommand{\arraystretch}{1.3}
\begin{tabular}{@{}lcccc@{}}
\toprule
\textbf{Method} & \textbf{Precomp.\ $N$-scaling} & \textbf{Inference $N$-scaling} & \textbf{\# PFN Forward Passes}\\
\midrule
Full Context & --- & $O(N^2)$ & $1$  \\
Uniform & $O(1)$ & $O(1)$ & $1$ \\
$k$NN (per query) & $O(N)$ & $O(1)$ & $T$  \\
MICP$^\dagger$ & $O(N^2)$ & $O(1)$ & $K^*$  \\
\textbf{CRUMB (ours)} & $O(N)$ & $O(1)$ & $K$ \\
\bottomrule
\end{tabular}
\end{table}

Once a context budget $n$ and the number of test samples $T$ are fixed, the PFN inference cost of all context selection methods becomes independent of $N$, and the only $N$-dependent costs are in precomputation (clustering, retrieval, or MMD herding). CRUMB's precomputation is $O(Nd(T + Kn))$, which is linear in $N$. Full-context inference, by contrast, is $O(N^2)$ and becomes prohibitively expensive for large $N$.

The distinction between CRUMB and per-query $k$NN is not in $N$-scaling, as both are linear, but in the number of PFN forward passes, which in practice dominates wall-clock time since each pass involves a full transformer forward pass through $L$ layers. Per-query $k$NN requires $T$ separate forward passes (one per test point), because potentially no two test points share a context. CRUMB requires only $K$ passes (one per test cluster), because all test points within a cluster share the same training context and can be batched together. For $T = 5{,}000$ test points and $K = 20$ clusters, this is a $250\times$ reduction in the number of forward passes. 

We provide details of the computational costs for each method compared in this paper. Let $N$ be the training set size, $T$ the test set size, $d$ the feature dimension, $n$ the context budget per forward pass (capped at the model window $N_{\max}$), $K$ the number of test clusters (CRUMB), and $K^*$ number of training clusters with test points routed to them (MICP).

\textbf{PFN Standard Inference.} The standard PFN takes the full training set ($N$ points) and all test queries ($T$ points) as input. In all three architectures we consider (TabPFNv2, TabICLv1, TabICLv2), the attention pattern is \emph{asymmetric}: training tokens attend to all other training tokens (self-attention), while test tokens attend only to training tokens (cross-attention). Test tokens do \emph{not} attend to each other. While the architectures may have different scalings overall, when considering this self-attention and cross-attention over the samples, the attention cost per layer can be broken down into: (a) train-to-train self-attention cost of $O(N^2)$, and (b) test-to-train cross-attention cost of $O(TN)$. Hence any architecture that employs this style of attention across samples has a cost that scales as $C_{\text{attn}} = O\bigl(N^2 + TN \bigr)$. The quadratic scaling in the number of training samples motivates the need for context selection methods, in which we can use $n < N$ samples instead.

\textbf{Full Context.} In selecting the full context, we incur a $O(1)$ selection cost, as no pre-processing is required, and a PFN inference cost, where the attention cost will scale as $O(N^2 + NT)$.

\textbf{Uniform Subsampling.} Draw $n$ training points uniformly at random and use them as context for all $T$ test queries in a single forward pass.
This breaks down into a selection cost of $O(n)$ for random index generation, and 
a PFN inference cost, where the attention cost will scale as $O(n^2 + nT)$.


\newpage

\textbf{Per-Query $k$NN Selection.}
For each test point $\bm{x}_j^*$ individually, retrieve its $n$ nearest neighbours from $\mathcal{D}_{\text{train}}$ and run a separate PFN forward pass with context $\mathcal{N}_j$ and a single test query. The procedure is summarised in Algorithm~\ref{alg:knn}.

\begin{algorithm}[H]
\caption{Per-Query $k$NN Context Selection for PFN Inference}
\label{alg:knn}
\begin{algorithmic}[1]
\REQUIRE Training set $\mathcal{D}_{\text{train}} = \{(\bm{x}_i, y_i)\}_{i=1}^{N}$, test set $\mathcal{D}_{\text{test}} = \{\bm{x}_j^*\}_{j=1}^{T}$, number of neighbours $n$, PFN model $f$
\STATE Standardise features of $\mathcal{D}_{\text{train}}$ and $\mathcal{D}_{\text{test}}$
\STATE Compute pairwise distances between $\mathcal{D}_{\text{test}}$ and $\mathcal{D}_{\text{train}}$
\FOR{$j = 1, \dots, T$}
    \STATE $\mathcal{N}_j \leftarrow n\text{-nearest-neighbours}(\bm{x}_j^*, \mathcal{D}_{\text{train}})$ \hfill $\triangleright$ Retrieve $n$ neighbours
    \STATE $\hat{y}_j \leftarrow f(\mathcal{N}_j, \{\bm{x}_j^*\})$ \hfill $\triangleright$ PFN forward pass (single test point)
\ENDFOR
\RETURN $\{\hat{y}_j\}_{j=1}^{T}$
\end{algorithmic}
\end{algorithm}

For finding the $n$-nearest neighbours of each of $T$ test points, we need to compute the distance to all $N$ training points and select the $n$ closest. 
Each distance computation costs $O(d)$ (where $d$ is the dimension of the data), giving $O(Nd)$ per query and a total cost of $C_{\text{retrieve}} = O(TNd)$.
We remark that this pre-computation cost for per-query $k$NN can be improved by using standard techniques like ball tree search~\cite{omohundro1989five}, where one can build a ball tree data structure in $O(N d (\log(N))^2)$ time. In the literature, the query complexity of ball tree search for $n$ nearest neighbours is sometimes claimed to be $O(n d \log(N))$ (which we call ``typical cost" for simplicity), for example in MICP~\cite{xu2024mixtureincontextprompterstabular}, while the worst case complexity is given as $O(N n d)$ (see Table~3 in~\cite{abbasifard2014survey}). Thus, the typical cost can be improved to $O(N d (\log(N))^2 + T n d \log(N))$. Note that the details of complexity calculation depend on the exact algorithm for tree search being implemented. While the lookup cost can be improved, this still does not allow batching in general, requiring $T$ forward passes.

Since each test point has a unique PFN instance being run separately (no batching), and for a single test point $T=1$, a forward pass for PFN inference costs $O(n^2 + n)$, running this for every test point gives a final inference cost of $O(Tn^2 + Tn)$.

\textbf{MICP.} MixturePFN~\cite{xu2024mixtureincontextprompterstabular} introduces a Sparse Mixture of In-Context Prompters (MICP). It clusters the \emph{training} data into $K'$ clusters, constructs a fixed-size ``prompt'' (support set) of $n$ training points per cluster, and routes each test point to the nearest cluster centroid at inference time.

A critical design choice is that the \textit{number of prompters} grows with $N$ as
\begin{equation}\label{eq:mixpfn_K}
    K' = \left\lceil \frac{\gamma N}{n} \right\rceil,
\end{equation}
where $\gamma > 0$ is a hyperparameter trading efficiency for effectiveness. Since $n$ is fixed (the context budget), $K'$ grows \emph{linearly} with $N$.
MICP incurs a one-time precomputation cost at initialization.

The MICP algorithm is composed of three steps: (a) clustering the training data, (b) building a ball tree over the $K'$ cluster centroids and $N$ training data points, and (c) constructing the prompter support sets onto which we route the test points. We will compute the cost associated with each step below.

In computing the clusters, we incur the $k$-means cost of using Lloyd's algorithm to cluster $N$ training points into $K' = \lceil \gamma N / n \rceil$ clusters. With $I$ iterations this cost is quantified as
\[
    C_{\text{cluster}} = O(NK'dI) = O\!\left(\frac{\gamma I N^2 d}{n}\right).
\]
Since $K' \propto N$, this scales quadratically in $N$.

After the clusters have been built, the ball tree construction over the $K'$ cluster centroids (for efficient routing) and over the full training data (for $k$NN expansion of small clusters) has cost~\cite{xu2024mixtureincontextprompterstabular}
\[
    C_{\text{tree}} = O(K' d \log(K') + Nd\log N) = O(N d \log(N)).
\]
Note that we use the ball tree construction cost given in~\cite{xu2024mixtureincontextprompterstabular} for consistency, instead of $O(N d (\log(N))^2)$ cost given in other studies~\cite{abbasifard2014survey}. We keep track of the dimension unlike~\cite{xu2024mixtureincontextprompterstabular}.

Finally, for each of the $K'$ prompters, we form a support set of size $n$. If the cluster has $\geq n$ points, we subsample, and if it has $< n$ points, we expand via $n$-nearest-neighbour search from the centroid. In the worst case, where $O(K')$ clusters need to be expanded, this has cost given as
\[
    C_{\text{prompt}} = O(K' n d\log N) = O\!\left(\frac{\gamma N}{n} \cdot nd \log N\right) = O(\gamma Nd \log N),
\]
using the fact that the ball tree search typically takes $O(n d \log(N))$ time  for finding $n$-nearest neighbours.
We remark that in the worst case, ball tree look up costs $O(N n d)$~\cite{abbasifard2014survey}.

The total pre-computation cost then amounts to~\cite{xu2024mixtureincontextprompterstabular}
\begin{equation}\label{eq:mixpfn_precomp}
    T_{\text{MixPFN, init}} = O\!\left(\frac{\gamma I N^2 d}{n} + Nd\log N\right).
\end{equation}

At inference, each test point is routed to its nearest cluster centroid via ball-tree search over the $K'$ centroids. 
This has cost:
\[
    C_{\text{route}} = O(T \log K') = O(T \log(N/n)).
\]
The routing step is $O(\log N)$ per test sample.

Finally, we run PFN inference. Let $K^* \in \{1, \dotsc, K'\}$ denote the number of unique clusters that test points are assigned to. Then, in the worst case, each such cluster is assigned $T / K^*$ test points, and correspondingly, the PFN inference for $K^*$ forward passes have a total cost of $O(K^* (n^2 + n (T/K^*))) = O(K^* n^2 + n T)$. Note that if $T$ is comparable to $N$, then $K^* \approx K' = O(N)$ in the worst case, and therefore, the number of forward passes for MICP can be large in principle.

\textbf{CRUMB.} CRUMB has three stages: (1)~cluster the test queries, (2)~greedy MMD selection per cluster, (3)~batched PFN inference.

Clustering the $T$ test points via $k$-means (Lloyd's algorithm) into $K$ clusters in $\mathbb{R}^d$ has cost
\[
    C_{\text{Stage 1}} = O(TKdI).
\]
The next step is greedy MMD selection: for each cluster $C_k$ (with $|C_k|$ test points), the greedy kernel herding procedure selects $n$ training points from the full pool of $N$ candidates.
First, we pre-compute cross-terms, where for each of $N$ training points, the computation of the mean kernel similarity to all $|C_k|$ test points in the cluster (first term in Eq.~\eqref{eq:greedy_score}) incurs the cost $O(N |C_k| \text{cost}(\kappa))$, where $\text{cost}(\kappa)$ is the cost of computing the kernel value given two points. For RBF kernel, we have $\textnormal{cost}(\kappa) = O(d)$ corresponding to computing $\|\bm{x}_i - \bm{x}_j^*\|^2$ and the exponential.
Thus, for RBF kernel used in this study, the associated cost per cluster for cross-terms is $C_{\text{cross}} = O(N \cdot |C_k| \cdot d)$.

Next, we compute the cost for the second term in Eq.~\eqref{eq:greedy_score} in the greedy loop.
This involves keeping track of the kernel costs between a growing batch $\mathcal{S}_k$ and the remaining training data.
Specifically, at time $t$, this amounts to a cost of $O((N - t) (t + 1) d)$ for the RBF kernel.
Over $n = |\mathcal{S}_k|$ steps, we pay a total cost of $C_{\textnormal{greedy}} = \sum_{t = 0}^{n - 1} O((N - t) (t + 1) d) = O(N n d)$.
Importantly, since we have already precomputed the cross terms, they do not need to be computed at every step.

As a result, the total cost for one cluster $C_k$ is $C_{\text{Stage 2}}^{(k)} = O\bigl(N|C_k|d + nNd\bigr) = O\bigl(Nd(|C_k| + n)\bigr)$.
Aggregating over all $K$ clusters, we obtain
\begin{equation}\label{eq:crumb_mmd_total}
    C_{\text{Stage 2}} = \sum_{k=1}^K O\bigl(Nd(|C_k| + n)\bigr) = O\bigl(Nd\bigl(\underbrace{\textstyle\sum_k |C_k|}_{= T} + Kn\bigr)\bigr) = O\bigl(Nd(T + Kn)\bigr).
\end{equation}
With $n$ capped at $N_{\max}$ and $K, T, d$ fixed, this is $O(N)$, linear in the training pool size.

Now, we look at the final state of batched PFN inference.
Here, we are required to run $K$ forward passes. Pass $k$ has context $\mathcal{S}_k$ (size $n$) and test batch $C_k$ (size $|C_k|$). For a single cluster the attention cost across samples for a single forward pass is $O(n^2 + |C_k|n)$. Therefore, running $K$ total forward passes gives a cost of $C_{\textnormal{Stage 3}} = O(Kn^2 + \sum_k |C_k|n) = O(Kn^2 + Tn)$.

As shown in the previous examples we can see that regardless of how the $T$ test points are batched, the cross-attention contribution in the attention always scales $O(Tn)$. However, as each of the $K$ batches uses a different training context the self attention for all batches scales as $O(Kn^2)$. For per-query $k$NN we get the worst case scaling of $O(Tn^2)$, which motivates CRUMB as a faster method that still preserves performance. We also note that better methods of finding training contexts can lead to a smaller context being needed for the same result. This way speedups can be achieved by making $n$ smaller.

\end{document}